\documentclass[lettersize,journal]{IEEEtran}
\usepackage{amsmath,amsfonts}
\usepackage{algorithmic}
\usepackage{algorithm}
\usepackage{array}
\usepackage[hidelinks]{hyperref}
\usepackage[caption=false,font=normalsize,labelfont=sf,textfont=sf]{subfig}
\usepackage{textcomp}
\usepackage{stfloats}
\usepackage{url}
\usepackage{verbatim}
\usepackage{graphicx}
\usepackage{cite}
\usepackage{arydshln}
\usepackage{tabularx}
\usepackage{multirow}
\usepackage{booktabs}  
\usepackage{xcolor}

\begin{document}

\title{Quantization Meets Spikes: Nearly Lossless Conversion at the First Timestep via Polarity Multi-Spike Mapping}

\author{
Hangming~Zhang\IEEEauthorrefmark{2},
Zheng~Li\IEEEauthorrefmark{2},
Chenxiang~Ma,
Huajin~Tang,~\IEEEmembership{Fellow,~IEEE},
Long~Cheng,~\IEEEmembership{Fellow,~IEEE},
Kay~Chen~Tan,~\IEEEmembership{Fellow,~IEEE},
and Qiang~Yu\IEEEauthorrefmark{1},~\IEEEmembership{Senior~Member,~IEEE}

\thanks{Hangming~Zhang and Qiang~Yu are with the School of Artificial Intelligence, Tianjin University, Tianjin 300350, China.}
\thanks{Zheng~Li is with the Tianjin International Engineering Institute, Tianjin University, Tianjin 300350, China.}
\thanks{Chenxiang~Ma and Kay~Chen~Tan are with the Department of Data Science and Artificial Intelligence, The Hong Kong Polytechnic University, Hong Kong SAR.}
\thanks{Huajin~Tang is with College of Computer Science and Technology and the State Key Laboratory of Brain-Machine Intelligence, Zhejiang University, Hangzhou 310027, China.}
\thanks{Long~Cheng is with the School of Artificial Intelligence, University of Chinese Academy of Sciences, Beijing 100049, China, and also with the State Key Laboratory of Multimodel Artificial Intelligence Systems, Institute of Automation, Chinese Academy of Sciences, Beijing 100190, China.}
\thanks{\IEEEauthorrefmark{2}(Equal contribution: Hangming Zhang, Zheng Li.)}
\thanks{\IEEEauthorrefmark{1}(Corresponding author: Qiang Yu. Email: yuqiang@tju.edu.cn.)}
}



\maketitle

\begin{abstract}

Spiking neural networks (SNNs) offer advantages in computational efficiency via event-driven computing, compared to traditional artificial neural networks (ANNs). While direct training methods tackle the challenge of non-differentiable activation mechanisms in SNNs, they often suffer from high computational and energy costs during training. As a result, ANN-to-SNN conversion approach remains a valuable and practical alternative. These conversion-based methods aim to leverage the discrete output produced by the quantization layer to obtain SNNs with low latency. Although the theoretical minimum latency is one timestep, existing conversion methods have struggled to realize such ultra-low latency without accuracy loss. Moreover, current quantization approaches often discard negative-value information following batch normalization and are highly sensitive to the hyperparameter configuration, leading to degraded performance. In this work, we, for the first time, analyze the information loss introduced by quantization layers through the lens of information entropy. Building on our analysis, we introduce polarity multi-spike mapping (PMSM) framework and a hyperparameter initialization strategy tailored for the quantization layer. Our method achieves nearly lossless ANN-to-SNN conversion at the extremity, i.e., the first timestep, while also leveraging the temporal dynamics of SNNs across multiple timesteps to maintain stable performance on complex tasks. Extensive experiments on six image and neuromorphic datasets consistently demonstrate that PMSM achieves nearly lossless accuracy at the first timestep. Remarkably, despite operating under ultra-low-latency constraints, PMSM surpasses state-of-the-art direct training methods on multiple benchmarks. Additionally, PMSM reduces energy consumption by more than two orders of magnitude compared with ANNs and by over 55\% relative to existing ANN-to-SNN conversion methods. This work, therefore, provides an effective framework for constructing ultra-low-latency SNNs and paves the way for efficient real-time neuromorphic systems.
\end{abstract}


\section{Introduction}

\IEEEPARstart{S}{piking} neural networks (SNNs) are biologically inspired neural networks that encode and transmit information through discrete spike events. Compared with traditional artificial neural networks (ANNs), SNNs exhibit superior energy efficiency and neuromorphic hardware compatibility~\cite{jang2019introduction, zhang2020low, 10347028}. There are two mainstream paradigms for building deep SNNs: surrogate-gradient based direct training~\cite{bohte2000spikeprop, neftci2019surrogate, 10848017,11104926} and ANN-to-SNN based conversion~\cite{diehl2015fast, cao2015spiking,hu2023fast}. Direct training methods approximate non-differentiable impulse functions by unfolding the computational graph over time and using differentiable surrogate gradients. However, direct training methods typically incur high training costs due to the need for backpropagation through time, resulting in significant computational and memory overhead. In contrast, ANN-to-SNN conversion methods train ANNs using conventional backpropagation without temporally unrolling, then directly convert the pretrained weights to SNNs. These conversion-based methods enable efficient construction of deep SNNs while preserving the original model architecture, achieving performance comparable to ANNs on multiple large-scale datasets~\cite{Liu_Zhao_Chen_Wang_Jiang_2022}. In addition, due to the training efficiency and compatibility with existing architectures, ANN-to-SNN conversion methods are well-suited for deployment on resource-constrained neuromorphic hardware~\cite{10737095, 10976463}.

Although ANN-to-SNN conversion has shown advantages for constructing efficient SNNs, achieving nearly lossless conversion under ultra-low-latency constraints remains a fundamental challenge. Since ANN-to-SNN conversion involves bridging two inherently different computational paradigms, the problem should be understood from both the ANN and SNN sides. On the ANN side, to better match the firing-rate-based representation in SNNs, a common strategy is to quantize continuous-valued activations into discrete levels. However, to ensure non-negative firing rates, the quantization process typically discards negative activations, leading to inevitable information loss. On the SNN side, neurons encode and transmit information through discrete spikes, whose representational capacity within each timestep is severely limited, making it difficult to faithfully approximate continuous activations, particularly under ultra-low-latency settings. These challenges become even more severe in the single-timestep setting, where quantization-induced information loss and limited spike expressiveness jointly prevent accurate representation.

To achieve nearly lossless ANN-to-SNN mapping under the single-timestep constraint, prior studies have explored improvements from either the ANN side or the SNN side. On the ANN side, quantized activation functions~\cite{You_Xu_Nie_Deng_Guo_Wang_He_2024, Bu_Fang_Ding_Dai_Yu_Huang_2022, YANG2025107076, li2022quantization} discretize continuous activations into low-bit representations, thereby reducing the number of timesteps required for SNN inference. However, such approaches typically constrain activations to non-negative ranges, and lack an explicit mechanism to preserve the information distribution during quantization, resulting in significant information loss. On the SNN side, various approaches aim to enhance the representational capacity of spikes to better approximate ANN activations, including firing rate calibration~\cite{Deng_Gu_2021, Li_Deng_Dong_Gong_Gu, Li_Deng_Dong_Gu_2022}, multi-spike schemes~\cite{Song_Ma_Sun_Xu_Dang_Yu_2021,Lan_Zhang_Ma_Qu_Fu_2023,Hao_Shi_Liu_Yu_Huang_2024}, and polarity spikes~\cite{9328792, Yu_Ma_Song_Zhang_Dang_Tan_2022, 9308402} for encoding negative information. However, these strategies are primarily designed from the perspective of spike representation, lacking coordinated design with ANN quantization, and thus struggle to precisely align continuous activations with discrete spike representations within a single timestep.

To address the aforementioned issues, we propose polarity multi-spike mapping (PMSM) framework for ANN-to-SNN conversion, which jointly addresses the challenges from both the ANN and SNN sides. On the ANN side, to explicitly ensure information-preserving transmission during quantization, we analyze the problem from an information-theoretic perspective. We observe that the information entropy of activations is closely related to the quantization boundaries. Based on this insight, we propose an entropy-guided hyperparameter initialization strategy that determines the quantization bounds to achieve nearly lossless information preservation. On the SNN side, PMSM employs a polarity multi-spike mechanism to enhance the representational capacity of spikes within a single timestep, enabling accurate reconstruction of quantized activations. By jointly improving information preservation and spike expressiveness, PMSM effectively bridges the gap between continuous activations and discrete spike representations, enabling nearly lossless ANN-to-SNN conversion in the extreme first timestep case while maintaining stable performance across multiple timesteps. Extensive experiments demonstrate that PMSM achieves state-of-the-art performance with ultra-low latency across multiple architectures and datasets. In addition, our method significantly reduces energy compared with existing ANN-to-SNN conversion approaches, highlighting its potential for efficient real-time neuromorphic systems. The main contributions of this work are summarized as follows:

\begin{itemize}

    \item \textbf{Polarity multi-spike mapping framework:} 
    We propose a unified ANN-to-SNN conversion framework that explicitly enforces information-preserving mapping from ANN activations to SNN spike representations. The proposed PMSM jointly considers the quantization process on the ANN side and the spike generation mechanism on the SNN side, providing a solution for nearly lossless conversion under ultra-low-latency constraints.
    
    \item \textbf{Entropy-lossless polarity quantization activation:} 
    To address information loss induced by conventional quantized activation functions, especially the truncation of negative values, we first introduce a Polarity Quantized Activation (PQA) function that preserves both positive and negative activations. Building upon this design, we further develop a hyperparameter initialization strategy for quantization layers, which determines the quantization boundary from an information theory perspective, thereby achieving lossless information preservation.

    \item \textbf{Nearly lossless ANN-to-SNN conversion at the first timestep:} 
    To enable spiking neurons to convey rich information within a single timestep, we propose an Augmented Integrate-and-Fire (AIF) neuron that supports a polarity multi-spike firing mechanism. This design enables precise reconstruction of ANNs activations with ultra-low latency, achieving nearly lossless conversion at the first timestep.

    \item \textbf{Generalization across timesteps, architectures, and datasets:} 
    We evaluate the proposed framework across multiple timesteps, four representative architectures (VGG-16, ResNet-20, Graph Convolution Network and ViT-S), and both static and event-based datasets, including CIFAR-10, CIFAR-100, ImageNet, CIFAR10-DVS, N-Caltech101 and EvTouch-Objects. The proposed PMSM achieves state-of-the-art performance across all settings in this study.

    \item \textbf{Energy-efficient spiking inference:}
    We conduct a comprehensive energy analysis to quantify the computational cost of SNNs. Experimental results show that our method significantly reduces the number of spikes, leading to substantial energy savings compared to conventional quantized activation-based conversion methods. Specifically, our approach achieves more than 55\% energy reduction on both CIFAR-10 and CIFAR-100 at $T=1$ and $T=2$, highlighting its potential for ultra-low-power neuromorphic systems.

\end{itemize}

\section{Related Works}

\subsection{ANN-to-SNN Conversion}

ANN-to-SNN conversion is a dominant paradigm for constructing high-performance SNNs with low latency. Existing methods primarily aim to reduce conversion errors through two representative directions: parameter calibration and activation function design.

Parameter calibration methods focus on minimizing the discrepancy between ANN activations and SNN firing behaviors by carefully adjusting network parameters. Early studies introduced SNN-adapted modules and provided theoretical analysis of conversion errors~\cite{Rueckauer_Lungu_Hu_Pfeiffer_Liu_2017}. Subsequent works demonstrated that appropriate membrane potential initialization can achieve nearly lossless conversion~\cite{Bu_Ding_Yu_Huang_2022}, and further improved performance by optimizing both initial and residual membrane potentials~\cite{Hao_Bu_Ding_Huang_Yu_2023, Hao_Ding_Bu_Huang_Yu_2023}. Other approaches incorporate threshold balancing, soft reset mechanisms, and second-order analysis to reduce activation mismatch via layer-wise calibration~\cite{Deng_Gu_2021, Li_Deng_Dong_Gong_Gu, Li_Deng_Dong_Gu_2022}.  Despite their effectiveness, calibration-based methods typically rely on layer-wise tuning and architecture-specific adjustments, which increase implementation complexity and limit scalability to diverse network structures.

Activation function design provides an alternative strategy by modifying ANN activations to better match spike-based representations. Continuous-valued activation functions~\cite{Ding_Yu_Tian_Huang_2021, Wang_Cao_Chen_Feng_Wang_2023, Han_Wang_Shen_Tang_2023, Jiang_Anumasa_Masi_Xiong_Gu_2023} and quantized activation schemes~\cite{Bu_Fang_Ding_Dai_Yu_Huang_2022} have been proposed to discretize activations, thereby reducing the number of timesteps required for SNN inference. These approaches enable efficient conversion of large-scale architectures, including Transformer-based models~\cite{Wang_Fang_Cao_Zhang_Wang_Xu_2023, You_Xu_Nie_Deng_Guo_Wang_He_2024, Hwang_Lee_Park_Lee_Kung_2024}.  However, most activation-based methods restrict activations to non-negative ranges, thereby discarding negative information and distorting the activation distribution. Moreover, they lack an explicit mechanism to preserve information during quantization, which becomes a critical limitation under ultra-low-latency settings.

In contrast to these approaches, our method explicitly targets the preservation of activation distributions during quantization, thereby minimizing information loss while jointly modeling polarity and magnitude.
\subsection{Spiking Neural Models in Conversion Process}

The choice of spiking neuron model plays a crucial role in determining the representational capacity of converted SNNs. Integrate-and-Fire (IF) neurons, which accumulate input currents without leakage, are widely adopted due to their ability to approximate ANN activations~\cite{Han_Srinivasan_Roy_2020}. However, the binary firing mechanism of standard IF neurons inherently limits their expressiveness within each timestep. To address this issue, compensation mechanisms have been proposed to better align firing rates with ANN activations~\cite{liuefficient}, although such methods still rely on rate-based approximations over multiple timesteps.

To improve the instantaneous representational capacity of SNNs, Song et al. first introduced the concept of augmented spikes~\cite{Song_Ma_Sun_Xu_Dang_Yu_2021, 9328792}, which extends the conventional binary spike representation by allowing multiple spikes to be emitted within a single timestep. This mechanism significantly enhances the information transmission capability of neurons while maintaining event-driven computation. Building upon this idea, a series of multi-spike neuron models have subsequently been proposed~\cite{Lan_Zhang_Ma_Qu_Fu_2023, Hao_Shi_Liu_Yu_Huang_2024}, further improving the instantaneous expressiveness of SNNs. In parallel, polarity-based neuron models have been developed~\cite{Wang_Zhang_Chen_Qu_2022, Wang_Liu_Zhang_Luo_Qu_2024, Guo_Chen_Liu_Peng_Zhang_Huang_Ma_2024, 10672817, Yu_Ma_Song_Zhang_Dang_Tan_2022, 9308402}, enabling neurons to emit both positive and negative spikes to better represent signed activations.

Despite these advances, existing neuron designs primarily focus on enhancing spike expressiveness from the SNN side, without coordinated consideration of how ANN activations are quantized. In particular, the loss of polarity information during ANN quantization is not explicitly addressed at the neuron level, leading to a mismatch between the quantized activations and spike responses, especially under single-timestep constraints.

In this work, we address this limitation through a joint design of activation quantization and spike generation. Specifically, we propose a PQA function to preserve both magnitude and sign information during quantization, together with an AIF neuron that employs a polarity multi-spike mechanism, enabling accurate reconstruction of both positive and negative activations within a single timestep.

\section{Limitations of Quantized Activation in ANN-to-SNN Conversion}
In this section, we analyze the fundamental limitations of quantized activation (QA) in existing ANN-to-SNN conversion frameworks. We first review the standard QA-based mapping mechanism from ANN activations to SNN firing behavior. We then provide an information-theoretic analysis to reveal the inherent entropy loss introduced by such quantization schemes, which limits the information preservation capability during conversion.
\subsection{Quantized Activation-Based ANN-to-SNN Conversion}
ANN-to-SNN conversion aims to transform a trained ANN into an SNN with the same architecture. The core idea is to approximate the continuous-valued activations in the ANN using the firing rates of spiking neurons in the SNN.
Among various neuron models, Integrate-and-Fire (IF) neurons are widely used in ANN-to-SNN conversion. Owing to their non-leaky membrane dynamics, the membrane potential accumulates input signals over time without decay~\cite{Han_Srinivasan_Roy_2020}, which enables spike firing rates to faithfully approximate ANN activations. Specially, the membrane potential dynamics of an IF neuron at layer $l$ are given by:
\begin{equation} \label{mem_acc}
    m^l(t) = v^l(t-1) + W^l_{\text{SNN}} s^{l-1}(t).
\end{equation}
Here, $m^l(t)$ and $v^l(t)$ denote the membrane potentials before and after spike firing, respectively. $W^l_{\text{SNN}}$ denotes the synaptic weights and $s^{l-1}(t)$ represents the input spikes from the previous layer. A spike is generated whenever the membrane potential exceeds a threshold $\vartheta^l_{\text{SNN}}$:
\begin{equation}
    s^l(t) = 
    \begin{cases}
        1, & m^l(t) \geq \vartheta^l_{\text{SNN}}, \\
        0, & \text{otherwise}.
    \end{cases}
\end{equation}
To preserve information during spike generation, a soft reset mechanism is typically employed:
\begin{equation} \label{soft_reset}
    v^l(t) = m^l(t) - \vartheta^l_{\text{SNN}} \cdot s^l(t),
\end{equation}
where only the portion of the membrane potential corresponding to the fired spikes is subtracted, allowing residual information to be retained for subsequent timesteps.

To this end, ANN activations are quantized into multiple levels using quantized activation (QA)~\cite{Bu_Fang_Ding_Dai_Yu_Huang_2022} to better match the discrete firing behavior of IF neurons. A conventional QA function is defined as:
\begin{equation}  \label{qa}
    y = f(x) =
    \vartheta \cdot \mathrm{clip}\left( \frac{1}{L} \left\lfloor \frac{x L}{\vartheta} \right\rfloor, 0, 1 \right),
\end{equation}
where $x = W^l a^{l-1}$ denotes the pre-activation input at layer $l$, $\vartheta$ is a learnable quantization threshold, and $L$ represents the number of quantization levels. It should be noted that $\vartheta$ is an auxiliary parameter of the ANN-side quantization function rather than the firing threshold of the spiking neuron in the converted SNN. During training, $\vartheta$ is jointly optimized with network weights through backpropagation. The output $y$ lies in a discrete set $\mathcal{Y} = \{0, \frac{\vartheta}{L}, \dots, \vartheta\}$.

Finally, to ensure consistency between the quantized ANN activations and the firing behavior of SNNs, a parameter mapping is introduced by scaling the ANN parameters as:
\begin{equation} \label{scaling}
    W^l_{\text{SNN}} = W^l \vartheta^l_{\text{SNN}}, 
    \quad
    \vartheta^l_{\text{SNN}} = \frac{\vartheta^l}{L}.
\end{equation}

Although this QA-based mapping provides an effective bridge between ANNs and SNNs, it fundamentally relies on non-negative quantization. Consequently, negative activations are inevitably discarded during the mapping process, leading to irreversible information loss.

\subsection{Entropy Loss in Quantized Activation}
To better analyze the information degradation introduced by quantized activation, we adopt an information-theoretic perspective. Specifically, we use information entropy $\mathcal{H}$ to measure the information capacity of activation distributions. The entropy of a discrete random variable $x$ is defined as:
\begin{equation}
\mathcal{H}(x) = -\sum_{i} p(x_i) \log p(x_i),
\end{equation}
where $x_i$ represents the $i$-th discrete value that $x$ can take, and $p(x_i)$ denotes the corresponding probability mass function. In practice, quantized activation (QA) is typically applied after batch normalization (BN). Therefore, we evaluate the information loss introduced by QA by comparing the entropy of BN outputs with that of their quantized counterparts.

We first analyze the entropy of BN outputs. BN layers normalize input features to have approximately zero mean and unit variance before the affine transformation. Although the subsequent learnable scaling and shifting parameters may alter the exact distribution, the normalized activations are commonly approximated as following a Gaussian-like distribution. Based on this approximation, the information entropy of BN outputs, denoted as $\mathcal{H}_{\text{BN}}$, can be expressed as:

\begin{equation}  \label{H_BN_1}
    \begin{aligned}
        \mathcal{H}_{\text{BN}}(x) =& - \int_{- \infty}^{+ \infty} \frac{1}{\sqrt{2 \pi}} e^{-\frac{x^2}{2}} \log(\frac{1}{\sqrt{2 \pi}} e^{-\frac{x^2}{2}}) \, dx  \\
        =& \frac{1}{2} \left( \int_{- \infty}^{+ \infty} \frac{1}{\sqrt{2 \pi}} e^{-\frac{x^2}{2}} \cdot \log 2\pi \, dx  \right. \\
        &  + \left. \quad \int_{- \infty}^{+ \infty} \frac{1}{\sqrt{2 \pi}} e^{-\frac{x^2}{2}} \cdot x^2 \, dx \right).
    \end{aligned}
\end{equation}
According to the definition of variance, $D(x) = E[x^2] - (E[x])^2$, the integral term in the above equation can be evaluated as follows:

\begin{equation}
    \int_{- \infty}^{+ \infty} \frac{1}{\sqrt{2 \pi}} e^{-\frac{x^2}{2}} \cdot x^2 \, dx= 1.
\end{equation}
Since BN outputs represent the normalized activation distributions that are directly fed into subsequent layers, they serve as the upper bound of information capacity:
\begin{equation}
    \mathcal{H}_{\text{BN}}(x) =  \frac{1}{2} (\log 2\pi e).
\end{equation}

We then analyze the information entropy of the QA output, which can be computed as follows:
\begin{equation}
    \mathcal{H}_{\text{QA}} = -\sum_{k=0}^LP(y_n) \log P(y_n).
\end{equation}
Here, $P(y_n)$ denotes the probability of output $y_n$. According to the definition of entropy, the entropy attains its maximum value when the output distribution is uniform over all possible outcomes. In this case, $P(y_n) = \frac{1}{L + 1}$, and the maximum entropy is given by $\mathcal{H}_\text{QA}^\text{max} = \log(L + 1)$. Theoretically, as $L \to \infty$, the entropy upper bound increases accordingly. However, due to the finite quantization threshold $\vartheta$, when the input $x \to \infty$, the output of the QA function becomes saturated at $\vartheta$, which means that $f(x) = \vartheta$. This saturation effect concentrates probability mass at the upper bound, thereby reducing entropy. Therefore, achieving maximal entropy requires both $L \to \infty$ and $\vartheta \to \infty$. Under these limiting conditions, Equation~\eqref{qa} simplifies to:
\begin{equation}
\lim_{\vartheta \to \infty} \lim_{L \to \infty} f(x) = \max(0, x) = \text{ReLU}(x).
\end{equation}

That is, the QA function becomes equivalent to the ReLU activation function when its information entropy reaches the maximum. Consequently, the output entropy of the ReLU function can be regarded as the theoretical upper bound of the information representation capacity of the QA function. When the input is assumed to approximately follow a Gaussian-like distribution with zero mean and unit variance, the probability density function (PDF) of the ReLU output can be expressed as:
\begin{equation} \label{relu_PDF}
    f_{\text{ReLU}}(x) = 
    \begin{cases}
        0,                                & x < 0, \\
        0.5 \cdot \delta(x),              & x = 0, \\
        \frac{1}{\sqrt{2\pi}} e^{-x^2/2}, & x > 0.
    \end{cases}
\end{equation}
In this context, $\delta(\cdot)$ denotes the Dirac delta function, representing a point mass centered at $x = 0$ with a probability weight of 0.5. Therefore, the output entropy of the ReLU function can be expressed as follows:
\begin{equation}  \label{H_ReLU_1}
\begin{aligned}
    \mathcal{H}_{\text{ReLU}}(x) =& -0.5 \log 0.5  -\int_0^\infty f_{\text{ReLU}}(x) \log f_{\text{ReLU}}(x)\, dx  \\
    =& 0.5 \log 2 -\int_0^\infty f_{\text{ReLU}}(x) \left( -\frac{1}{2} \log(2\pi) \right. - \frac{x^2}{2}) dx \\
    =&0.5 \log 2+\frac{1}{2} \log(2\pi) \cdot 0.5 + \frac{1}{2} \cdot 0.5\\
    \approx&0.69\mathcal{H}_{\text{BN}}(x).
\end{aligned}
\end{equation}

Thus, the information entropy of the ReLU output, when the input is approximated as following a standard normal distribution, is approximately $69\%$ of the entropy of the BN output. This value can also be regarded as the upper bound of the information representation capacity that the QA function can achieve under ideal conditions, namely, $L \to \infty$ and $\vartheta \to \infty$. Therefore,
\begin{equation}
\mathcal{H}_{\text{QA}}^{\text{max}} = \mathcal{H}_{\text{ReLU}} \approx 0.69 \mathcal{H}_{\text{BN}}.
\end{equation}
Our analytical result above reveals that conventional QA-based conversion inherently suffers from significant entropy loss, which fundamentally limits the amount of information that can be preserved during ANN-to-SNN conversion.
\begin{figure*}[t]  
  \centering
  \includegraphics[width=0.95\linewidth]{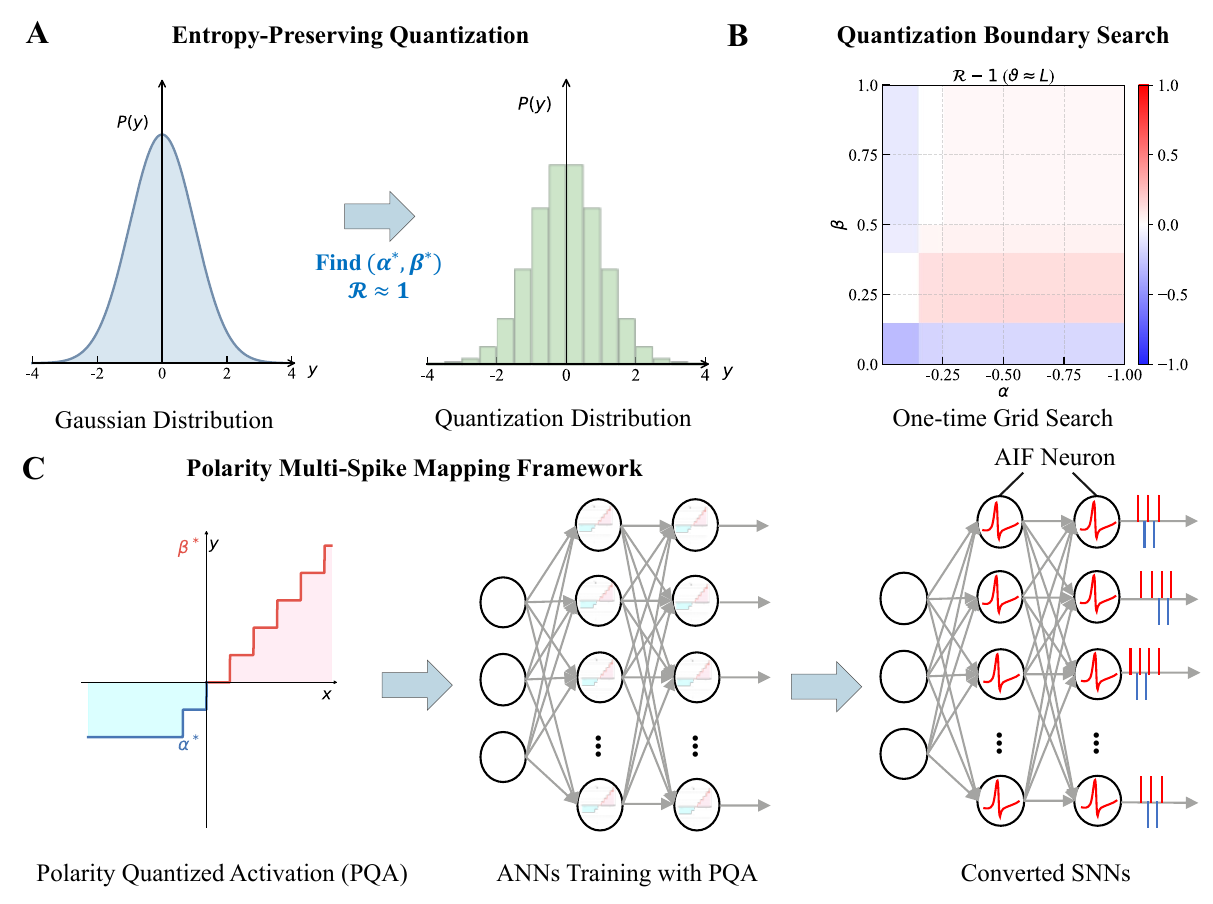}
  \caption{Overview of the proposed entropy-guided polarity mapping framework for ANN-to-SNN conversion. \textbf{(A)} Entropy-preserving quantization, where the original Gaussian-distributed activations are mapped to a quantized distribution by determining the optimal quantization bounds $(\alpha^*, \beta^*)$ such that the entropy ratio $\mathcal{R} \approx 1$. \textbf{(B)} The grid search results of $\mathcal{R} = \frac{\mathcal{H}_{\text{PQA}}}{\mathcal{H}_{\text{BN}}}$ under condition $L = \vartheta = 8$, with $\alpha \in [-1, 0]$ and $\beta \in (0, 1]$. The colormap visualizes the value of $\mathcal{R} - 1$:  Bluer regions indicate more information loss during transmission; white regions denote lossless transmission; redder regions reflect stronger noise injection during transmission. \textbf{(C)} Overall PMSM pipeline, where ANNs are first trained with the proposed polarity quantized activation (PQA), and then converted into SNNs through AIF neuron.}
  \label{fig:fig1}
\end{figure*}
\section{Methodology}

In this section, we present the polarity multi-spike mapping (PMSM) framework for nearly lossless ANN-to-SNN conversion under low-latency constraints, as illustrated in Fig.~\ref{fig:fig1}. The proposed framework consists of two key components. First, we introduce a polarity quantized activation (PQA) function and analyze its entropy to characterize information preservation during discretization, as shown in Fig.~\ref{fig:fig1}(A). In addition, we develop a hyperparameter initialization strategy for the quantization process, where the quantization boundaries are determined from an information-theoretic perspective to ensure lossless information preservation, as shown in Fig.~\ref{fig:fig1}(B). Next, as shown in  Fig.~\ref{fig:fig1}(C), the ANNs are trained with the proposed PQA to align activation distributions with spike-based representations. Then, we convert the trained ANNs into SNNs using an Augmented Integrate-and-Fire (AIF) neuron model. As illustrated in Fig.~\ref{fig:fig1}(C), the AIF neuron enables multi-spike firing with polarity within a single timestep, thereby enhancing the instantaneous representational capacity under extreme latency constraints. Finally, we conduct a theoretical analysis of conversion errors under both single- and multi-timestep settings to further validate the effectiveness of the proposed framework.

\subsection{Polarity Quantized Activation Function}  \label{section:3:1}
To address the information loss caused by truncating negative activations in conventional quantization, we propose a polarity quantized activation (PQA) function. The key idea is to explicitly introduce a quantized representation for negative values, thereby preserving the polarity information of activations. Unlike traditional quantization strategies that retain only non-negative responses, PQA encodes both positive and negative activations within a unified quantization framework. The formulation of PQA is given as follows:
\begin{equation}  \label{eq:PQA}
    y^l = \vartheta^l clip(
            \frac{1}{L} \lfloor \frac{x^l L}{\vartheta^l} \rceil,
            \alpha,
            \beta
        )
    ,\quad x^l = W^l y^{l-1}.
\end{equation}
Here, $y^{l-1}$ denotes the input to the $l$-th layer, while $W^l$ represents the corresponding weight matrix. $\vartheta^l$ is a learnable quantization threshold in the ANN-side PQA function, and $L$ specifies the number of quantization levels. Although $\vartheta^l$ is jointly optimized with network weights through backpropagation during training, its initialization and its relationship with $L$ are critical to information preservation and will be theoretically analyzed in the subsequent content. The operator $\lfloor \cdot \rceil$ denotes rounding. The parameters $\alpha$ and $\beta$ define the lower and upper bounds of the quantization range, subject to $-1 \leq \alpha \leq 0$ and $0 < \beta \leq 1$. This design ensures that the quantization range spans both negative and positive activations. Meanwhile, restricting the range within $[-1, 1]$ helps concentrate the quantization levels around the high-density region of the activation distribution.

To this end, we theoretically characterize the information preservation capability of PQA by deriving its entropy $\mathcal{H}_{\text{PQA}}$. Building on the PQA function, we define an intermediate variable $k = \left\lfloor \frac{xL}{\vartheta} \right\rceil$, where $k \in \mathbb{Z}$. For any integer $k$, from the inequality $k \leq \frac{xL}{\vartheta} < k + 1$, we can derive the corresponding interval $\frac{\vartheta}{L}k \leq x < \frac{\vartheta}{L}(k + 1)$. Since $x \sim \mathcal{N}(0, 1)$, $k$ approximately follows a normal distribution with mean $0$ and variance $\tfrac{L^2}{\vartheta^2}$. After scaling, the minimum and maximum values of k are given by $k_{\min} = L\alpha$ and $k_{\max} = L \beta$. The probability of $k$ is given by:
\begin{equation}  \label{eq:prob_k_Phi}
\begin{aligned}
    P(k=n) &= P\left( n - \tfrac{1}{2}\leq k < n+\tfrac{1}{2} \right)  \\
    &= \Phi\left( \frac{n+\tfrac{1}{2}}{L / \vartheta} \right)
    -
    \Phi\left( \frac{n-\tfrac{1}{2}}{L / \vartheta} \right).
\end{aligned}
\end{equation}
In this context, $\Phi(\cdot)$ denotes the cumulative distribution function (CDF) of the standard normal distribution.
Due to the clipping operation, the PMF of $y$ consists of three cases, depending on whether the input $x$ falls below $\vartheta \alpha$, within the range $[\vartheta \alpha, \vartheta \beta]$, or above $\vartheta \beta$.
Accordingly, the PMF of $y$ can be expressed as follows:
\begin{equation}
\begin{aligned}
    P_Y(y) =
    & \quad \delta (y - \vartheta \alpha) \sum_{k=-\infty}^{\lfloor L \alpha \rfloor - 1} p(k)  \\
    &+ \sum_{k = \lceil L\alpha \rceil}^{\lfloor L\beta\rfloor} \delta \left(y - \vartheta \frac{k}{L} \right) p(k)  \\
    &+ \delta(y - \vartheta \beta) \sum_{k= \lceil L \beta \rceil}^{+\infty} p(k).
\end{aligned}
\end{equation}
Here, $p(k) = \Phi\left( \frac{\vartheta}{L}(k + 1) \right) - \Phi\left( \frac{\vartheta}{L}k \right)$, where $\delta(\cdot)$ denotes the Dirac delta function, which captures the discrete probability mass at specific points.
Accordingly, the entropy of $y$ can be decomposed into three components: $\mathcal{H}_1$, $\mathcal{H}_2$, and $\mathcal{H}_3$, corresponding to the lower bound, intermediate region, and upper bound, respectively.

\begin{equation}  \label{eq:H_PQA}
    \begin{aligned}
        \mathcal{H}_{\text{PQA}} & = \mathcal{H}_1 + \mathcal{H}_2 + \mathcal{H}_3,                                                                                                                                            \\
        \mathcal{H}_1            & = - \Phi \left( \frac{k_{\mathrm{min}} - \tfrac{1}{2}}{L / \vartheta} \right) \log (\Phi \left( \frac{k_{\mathrm{min}} - \tfrac{1}{2}}{L / \vartheta} \right)), \\
        \mathcal{H}_2 &= - \sum_{k=k_{\mathrm{min}}}^{k=k_{\mathrm{max}}} \left[
    \left( \Phi \left( \frac{k + \frac{1}{2}}{L / \vartheta} \right) - \Phi \left( \frac{k - \frac{1}{2}}{L / \vartheta} \right) \right) \right. \\
    &\quad \cdot
    \left. \log \left( \Phi \left( \frac{k + \frac{1}{2}}{L / \vartheta} \right) - \Phi \left( \frac{k - \frac{1}{2}}{L / \vartheta} \right) \right)
\right], \\
        \mathcal{H}_3            & = - \left[1 - \Phi \left( \frac{k_{\mathrm{max}} - \frac{1}{2}}{L / \vartheta} \right)\right] \\
        &\quad \cdot
        \log \left(1 - \Phi \left( \frac{k_{\mathrm{max}} - \frac{1}{2}}{L / \vartheta} \right)\right).
    \end{aligned}
\end{equation}
The entropy terms $\mathcal{H}_1$, $\mathcal{H}_2$, and $\mathcal{H}_3$ are fundamentally determined by the ratio $\frac{\vartheta}{L}$, which governs the scaling of the discrete quantization variable $k$, thereby shaping the probability distribution of the PQA outputs and the resulting entropy $\mathcal{H}_{\mathrm{PQA}}$. To understand the information preservation capability of the PQA function under different quantization regimes, we analyze the behavior of $\mathcal{H}_{\mathrm{PQA}}$ under different scales of $\frac{\vartheta}{L}$.

\textbf{Case 1.} When $\vartheta \ll L$: The ratio $\frac{\vartheta}{L}$ becomes negligible. Consequently:
\begin{equation}  \label{eq:P_of_xx_case1}
    \Phi\left( \left( k - \tfrac{1}{2} \right) \cdot \frac{\vartheta}{L} \right) \approx \Phi(0) = 0.5.
\end{equation}
It follows from Equation~\eqref{eq:H_PQA} that:
\begin{equation}
    \mathcal{H}_1 = -0.5 \log0.5,\quad \mathcal{H}_2 = 0,\quad \mathcal{H}_3 = -0.5 \log0.5.
\end{equation}
Therefore, the entropy of the PQA outputs reduces to:
\begin{equation}
    \mathcal{H}_{\text{PQA}} = -\log 0.5.
\end{equation}
Using the previously derived entropy of BN outputs, the resulting entropy ratio is:
\begin{equation}
    \mathcal{R} = \frac{\mathcal{H}_{\text{PQA}}}{\mathcal{H}_{\text{BN}}}
    = \frac{-\log 0.5}{\frac{1}{2} \log(2\pi e)}
    \approx 0.49.
\end{equation}
This result indicates that when $\vartheta \ll L$, the quantization process suffers from severe entropy degradation, leading to substantial information loss.

\textbf{Case 2.} When $\vartheta \gg L$: The ratio $\frac{\vartheta}{L}$ becomes arbitrarily large. Hence:
\begin{equation}  \label{eq:P_of_xx_case2}
    \Phi \left( { \left( k - \tfrac{1}{2} \right) \cdot \frac{\vartheta}{L} } \right) \approx \Phi(\infty) = 0.
\end{equation}
Based on Equation~\eqref{eq:H_PQA}, we obtain:
\begin{equation}
    \mathcal{H}_1 = 0,\quad \mathcal{H}_2 = 0,\quad \mathcal{H}_3 = 0,
\end{equation}
which further leads to:
\begin{equation}
    \mathcal{H}_{\mathrm{PQA}} = 0.
\end{equation}
This result shows that when $\vartheta \gg L$, the PQA outputs collapse into an extremely sparse distribution with vanishing entropy, resulting in complete information degradation.

\textbf{Case 3.} Intermediate regime with comparable $\vartheta$ and $L$: 
The above analysis shows that the entropy of the PQA outputs collapses in both extreme regimes where $\vartheta \ll L$ and $\vartheta \gg L$. Therefore, effective information preservation requires $\vartheta$ and $L$ to remain within a comparable scale, preventing the entropy degradation caused by excessive imbalance between the quantization threshold and quantization levels.

Motivated by this observation, we initialize the quantization threshold according to
\begin{equation}
\vartheta = cL + \epsilon,\quad \epsilon \to 0,\quad 0.1 \leq c \leq 10,
\end{equation}
which ensures that $\vartheta$ and $L$ remain of the same order of magnitude. The range $0.1 \leq c \leq 10$ is considered as a practical search space. Since the entropy of the PQA function depends jointly on $(\vartheta,L,\alpha,\beta)$, exhaustive exploration of the full parameter space is computationally intractable. Moreover, Cases 1 and 2 have already shown that entropy preservation is impossible when $\vartheta$ and $L$ differ by several orders of magnitude. Therefore, we restrict the analysis to $0.1 \leq c \leq 10$, which captures the regime where $\vartheta$ and $L$ are of comparable scales and where entropy-preserving solutions are expected to exist. Under this condition, $\frac{\vartheta}{L} \approx c$, implying that the probability terms in Eq.~(18), i.e., $\Phi\left(\left(k-\frac{1}{2}\right)\frac{\vartheta}{L}\right)$, become primarily governed by the ratio $c$ rather than the individual values of $\vartheta$ and $L$. For analytical simplicity, we adopt $c=1$ in all subsequent experiments, yielding the initialization $\vartheta=L$. Under this condition, the entropy of the PQA outputs becomes primarily governed by the quantization bounds $(\alpha,\beta)$. As illustrated in Fig.~\ref{fig:fig1}(B), there exist regions where the entropy ratio $\mathcal{R} = \frac{\mathcal{H}_{\text{PQA}}}{\mathcal{H}_{\text{BN}}} \approx 1$, indicating lossless information preservation. This observation suggests that, under initialization of $\vartheta = L$, appropriate choices of $(\alpha, \beta)$ can align quantization intervals with high-density regions of the input distribution, thereby preserving the full information content. It should be noted that the above condition serves only as a theoretically motivated initialization strategy and does not constrain the optimization process during training.

In summary, the relationship between the quantization threshold $\vartheta$ and the quantization level $L$ plays a decisive role in the information preservation capability of the PQA function. Our analysis shows that when $\vartheta = cL + \epsilon$ with $\epsilon \to 0$, a set of feasible parameter configurations $(\alpha, \beta)$ can be identified to preserve the activation entropy without loss. In contrast, when $\vartheta$ and $L$ become severely imbalanced, the resulting mismatch inevitably leads to irreversible entropy degradation that cannot be compensated by any configuration of $(\alpha, \beta)$. Motivated by this observation, we formulate a  hyperparameter initialization strategy for the quantization layer. Specifically, the quantization threshold is initialized as $\vartheta=L$, and the corresponding feasible configuration $(\alpha,\beta)$ is selected from the entropy-preserving region identified by the above analysis. It is worth noting that the entropy analysis is used solely to determine the initialization of the quantization parameters. During network training, $\vartheta$ remains a learnable parameter and is further optimized through backpropagation together with the network weights, whereas the selected $(\alpha,\beta)$ are kept fixed throughout training. The resulting initialization strategy can be expressed as:
\begin{equation}
\begin{aligned}
\text{Find } (\alpha^*, \beta^*) \quad \text{s.t.}\quad
& \mathcal{R}(\alpha,\beta;\vartheta_0,L) \approx 1, \\
& \vartheta_0 = L.
\end{aligned}
\end{equation}
where $\mathcal{R} = \frac{\mathcal{H}_{\text{PQA}}}{\mathcal{H}_{\text{BN}}}$ denotes the entropy ratio between the quantized activation and the original activation distribution and $\vartheta_0$ denotes the initialization value of the learnable threshold. The constraint $\mathcal{R} \approx 1$ enforces lossless information preservation during initialization, while $\vartheta_0 = L$ ensures that the system starts from the entropy-preserving regime identified by the theoretical analysis. Nevertheless, the final objective of network optimization is classification performance rather than entropy preservation alone. Therefore, $\vartheta_0$ is not fixed after initialization and is subsequently optimized together with the network parameters through backpropagation to better adapt the quantization behavior to the downstream task. The solution $(\alpha^*, \beta^*)$ denotes a feasible set of quantization bounds that satisfy the specified lossless condition.
\begin{figure}[htb]
    \centering
    \includegraphics[width=\linewidth]{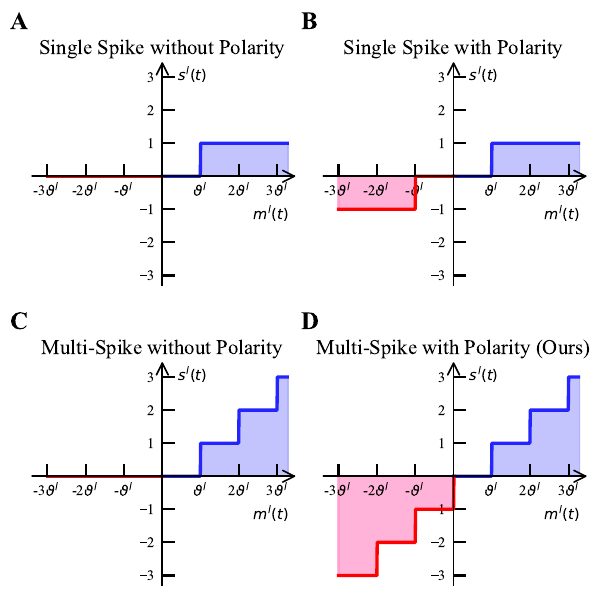} 
    \caption{Different spiking neuron firing mechanisms. The x-axis represents membrane potential, and the y-axis represents spike firing. The blue-shaded area indicates the positive spike region, while the red-shaded area represents the negative spike region. \textbf{(A)} and \textbf{(B)} correspond to non-polar and polar single-spike mechanisms, respectively, while \textbf{(C)} and \textbf{(D)} correspond to non-polar and polar multi-spike mechanisms.}
    \label{fig:fig3}
\end{figure}

\subsection{Augmented Multi-Spiking IF neurons}  \label{sec:AIF_neuron}
To preserve the entropy advantage of PQA during ANN-to-SNN conversion, the spike representation should faithfully encode both the magnitude and polarity information contained in the quantized activations. However, existing neuron models are unable to simultaneously satisfy these two requirements. As illustrated in Fig.~\ref{fig:fig3}(A), conventional single-spike neurons without polarity can only represent binary positive events\cite{Jiang_Anumasa_Masi_Xiong_Gu_2023}, resulting in the loss of both negative information and activation magnitude. Introducing polarity information, as shown in Fig.~\ref{fig:fig3}(B), alleviates the sign ambiguity but remains limited by the single-spike firing mechanism\cite{Wang_Liu_Zhang_Luo_Qu_2024}. Alternatively, multi-spike neurons without polarity, shown in Fig.~\ref{fig:fig3}(C), improve magnitude representation through spike counts but still discard negative-valued information\cite{Bu_Fang_Ding_Dai_Yu_Huang_2022}. Consequently, none of these designs can faithfully represent the discrete activations generated by PQA.

To address this issue, we propose an augmented integrate-and-fire (AIF) neuron model that enables neurons to generate multiple polarity spikes within a single timestep, as illustrated in Fig.~\ref{fig:fig3}(D). The key idea is to directly align spike generation with the discrete quantization behavior of PQA, thereby ensuring that spike counts faithfully reproduce the quantized activations at the first timestep and enabling accurate ANN-to-SNN conversion under ultra-low-latency constraints. 

Following the membrane potential accumulation process defined in Equations~\eqref{mem_acc}, the AIF neuron generates spikes according to both the magnitude and sign of the membrane potential $m^l(t)$. Specifically, when $m^l(t) > 0$, the neuron emits a number of positive spikes proportional to $\frac{m^l(t)}{\vartheta^l_{\text{SNN}}}$. When $m^l(t) < 0$, it emits negative spikes in a similar manner, reducing the membrane potential toward zero. To ensure stable spike generation, we further impose explicit upper and lower bounds on the number of spikes per timestep, denoted by $C_{\text{pos}}$ and $C_{\text{neg}}$, respectively. The output of the AIF neuron at timestep $t$ is thus defined as:
\begin{equation}  \label{eq:def:s} 
    s^{l}(t) = g\left(m^{l}(t)\right)
             = \mathrm{clip}
               \left(
                \left\lfloor \frac{m^l(t)}{\vartheta^l_{\text{SNN}}} \right\rfloor,
                C_{\text{neg}},
                C_{\text{pos}}
               \right),
\end{equation}
where $s^l(t) \in \mathbb{Z}$ represents the discrete count of binary spikes emitted at timestep $t$. A positive value indicates excitatory spikes, while a negative value indicates inhibitory spikes. The firing threshold $\vartheta^l_{\text{SNN}}$ is obtained from the ANN-to-SNN parameter mapping defined in Equation~\eqref{scaling}, ensuring consistency between the quantized ANN activations and the corresponding SNN firing behavior. The bounds are defined as $C_{\text{neg}} = \alpha^*L$ and $C_{\text{pos}} = \beta^*L$, which are directly derived from the PQA quantization range, ensuring consistency between ANN activations and SNN spike representations. To preserve information across timesteps, we adopt the soft-reset mechanism proposed in~\cite{Han_Srinivasan_Roy_2020}. After spike generation, the membrane potential is updated according to the soft-reset rule defined in Equation~\eqref{soft_reset}.

In summary, the proposed AIF neuron establishes a direct correspondence between spike counts and the discrete activations defined by PQA. By introducing explicit constraints on spike generation within each timestep, it enables precise control over the emitted spikes while maintaining stable dynamics. This design ensures that the quantized activations can be faithfully reconstructed in the SNN side, thereby supporting nearly lossless ANN-to-SNN conversion at the first timestep.


\subsection{Conversion Error Analysis} \label{appx:mapping_error}
In this subsection, we analyze the conversion error of the proposed framework under different timestep conditions. Specifically, we consider both the single-timestep case ($T=1$) and the multi-timestep case ($T>1$). In ANN-to-SNN conversion, the input to each layer remains constant across timesteps, as the same weighted input is repeatedly accumulated. Accordingly, the SNNs at layer $l$ is driven by a constant weighted input $z^l$ at each timestep. Based on this formulation, the approximate form of the conversion error can be expressed as:
\begin{equation}  \label{eq:def:widetide_Err_apx}
   \widetilde{{Err}^l}
       = \frac{1}{T} \sum_{i = 1}^{T} \left( \left\lfloor \frac{v^l(i-1)+ z^l}{\vartheta^l_{\text{SNN}}} \right\rfloor \right)
       -
        \left\lfloor \frac{z^l }{\vartheta^l_{\text{SNN}}} \right\rceil.
\end{equation}

We first analyze the single-timestep case ($T = 1$). When the initial membrane potential satisfies $\lim_{\epsilon \to 0} v^l(0) = \frac{1}{2}\vartheta^l_{\text{SNN}} + \epsilon$, the Equation~\eqref{eq:def:widetide_Err_apx} can be simplified to:
\begin{equation}
\begin{aligned}
\left. \widetilde{\mathrm{Err}^l} \right|_{T=1}
&\approx
\left\lfloor \frac{v^l(0)+ z^l}{\vartheta^l_{\text{SNN}}} \right\rfloor
-
\left\lfloor \frac{z^l }{\vartheta^l_{\text{SNN}}} \right\rceil \\
&\approx
\left\lfloor \frac{\tfrac{1}{2}\vartheta^l_{\text{SNN}}+\epsilon + z^l}{\vartheta^l_{\text{SNN}}} \right\rfloor
-
\left\lfloor \frac{z^l }{\vartheta^l_{\text{SNN}}} \right\rceil
\approx 0.
\end{aligned}
\end{equation}
This result indicates that, with proper initialization of the membrane potential similar as approaches in \cite{Bu_Fang_Ding_Dai_Yu_Huang_2022}, the SNNs can accurately reproduce the corresponding quantized activation within a single timestep, resulting in negligible conversion error.

We then consider the multi-timestep case ($T > 1$). Based on the membrane potential update mechanism of the SNNs, we have:
\begin{equation}
    v^l(i) = v^l(i-1) + z^l - \vartheta^l_{\text{SNN}} \cdot s^l(i-1).
\end{equation}
Here, $s^l(i) = \left\lfloor \frac{v^l(i-1) + z^l}{\vartheta^l_{\text{SNN}}} \right\rfloor$ (according to Equation \eqref{eq:def:s}) denotes the number of spikes emitted at the $i$-th timestep.
Given that the initial membrane potential satisfies $v^l(0) = \tfrac{1}{2} \vartheta^l_{\text{SNN}} + \epsilon$ and the input $z^l$ remains constant, the soft-reset mechanism ensures that $v^l(i)$ remains bounded within the range $(-\vartheta^l_{\text{SNN}}, \vartheta^l_{\text{SNN}})$ for all $i$.
The spike function can thus be further expressed as follows:
\begin{equation}
    s^l(i) = \lfloor \frac{v^l(i-1) + z^l}{\vartheta_{\text{SNN}}} \rfloor = \lfloor \frac{v^l(0) + z^l} {\vartheta_{\text{SNN}}} + \frac{\delta_i}{\vartheta_{\text{SNN}}} \rfloor .
\end{equation}
Let $\delta_i = v^l(i{-}1) - v^l(0)$ denote the perturbation introduced by the deviation of the membrane potential at timestep $i-1$ from its initial value.
The error term can then be further approximated as:
\begin{equation}
\begin{aligned}
&\sum_{i = 1}^{T}  \left\lfloor \frac{v^l(0) + z^l} {\vartheta_{\text{SNN}}} + \frac{\delta_i}{\vartheta_{\text{SNN}}} \right\rfloor \\
=
& T \cdot \left\lfloor \frac{z^l}{\vartheta^l_{\text{SNN}}} + \tfrac{1}{2} + o(\epsilon) \right\rfloor + \sum_{i = 1}^{T}\Delta_i,
\end{aligned}
\end{equation}
where $\Delta_i=\left\lfloor\frac{v^l(0)+z^l}{\vartheta_{\text{SNN}}}+\frac{\delta_i}{\vartheta_{\text{SNN}}}\right\rfloor-\left\lfloor\frac{v^l(0)+z^l}{\vartheta_{\text{SNN}}}\right\rfloor,\quad\Delta_i\in\{-1,0,1\}$, and by substituting this into Equation~\eqref{eq:def:widetide_Err_apx}, we obtain:
\begin{equation}
\begin{aligned}
\left. \widetilde{\mathrm{Err}}^l_{\mathrm{apx}} \right|_{T>1}
&\approx \frac{1}{T}\cdot T \left( \left\lfloor \frac{z^l}{\vartheta^l_{\mathrm{SNN}}}  + \frac{1}{2} + o(\epsilon) \right\rfloor \right) \\
& \quad + \frac{1}{T}\sum_{i = 1}^{T}\Delta_i
- \left\lfloor \frac{z^l}{\vartheta^l_{\mathrm{SNN}}} \right\rceil \\
      &=  \left\lfloor \frac{z^l }{\vartheta^l_{\mathrm{SNN}}} + o(\epsilon)  \right\rceil  \\
         &\quad -
         \left\lfloor \frac{z^l }{\vartheta^l_{\mathrm{SNN}}} \right\rceil + \frac{1}{T}\sum_{i = 1}^{T}\Delta_i \\
      &= \frac{1}{T}\sum_{i = 1}^{T}\Delta_i.
\end{aligned}
\end{equation}
\begin{figure}[htb]
    \centering
    \includegraphics[width=\linewidth]{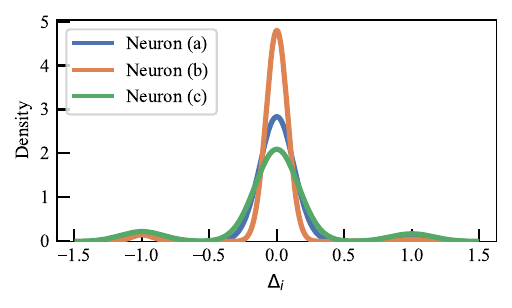}
    \caption{Kernel density estimates of the spike-count difference $\Delta_i$ between each timestep and the first timestep across 100 timesteps from three randomly selected neurons in VGG-16.}
    \label{fig:figappx2}
\end{figure}

This expression indicates that, in the multi-timestep case ($T > 1$), the conversion error is caused by the difference in spike count $\Delta_i$ at each timestep $i$, as illustrated in Fig.~\ref{fig:figappx2}. Since $\Delta_i \in \{-1, 0, +1\}$, the deviation at each timestep is bounded. Let $\mathbb{D} = \{ \Delta_i \mid i \in [1, T] \}$ denote the set of spike count deviations across timesteps. The overall conversion error is characterized by the average deviation $\frac{1}{T} \sum_{i=1}^{T} \Delta_i$. As $T \to \infty$, the influence of individual deviations is progressively averaged out, and the empirical mean $\frac{1}{T} \sum_{i=1}^{T} \Delta_i$ tends to approach zero. This indicates that the accumulated deviation does not grow unbounded with time, but instead remains controlled. Therefore, as the number of timesteps increases, the impact of spike count deviations on the ANN-to-SNN conversion accuracy becomes increasingly negligible.

\section{Experiments}
This section presents a comprehensive evaluation of the proposed method across diverse architectures and datasets. Section V-A presents the experimental setup. Section V-B compares the proposed method with both state-of-the-art ANN-to-SNN conversion approaches and representative direct training methods to validate its effectiveness. Section V-C presents a layer-wise entropy ratio analysis across different architectures, in order to provide empirical evidence that the proposed initialization strategy consistently guides training toward information-preserving quantization configurations. Section V-D investigates the hyperparameter initialization strategy, analyzing its impact on representation capability and conversion accuracy. Finally, Section V-E evaluates the energy efficiency of the proposed framework.
\renewcommand{\arraystretch}{1.25}
\begin{table}[!htb]
    \centering
    \small
    \caption{Training hyperparameters used for VGG-16 and ResNet-20 on CIFAR-10 and CIFAR-100.\label{tabel:appxtabel:vgg_and_resnet}}
    \label{tab:argparse_config}
    \begin{tabularx}{\linewidth}{Xc}
    \toprule
    \textbf{Hyperparameter} & \textbf{Value} \\
    \midrule
    Batch Size & 300 \\
    Total Train Epoch & 600 \\
    Initial Learning Rate & 1e-1 \\
    Weight Decay & 5e-4 \\
    $L$ & 8 \\
    ANN Threshold $\vartheta$ & 8 \\
    $\alpha^*$ & -0.25 \\
    $\beta^*$ & 1 \\
    \bottomrule
\end{tabularx}
\end{table}


\begin{table}[!htb]
    \centering
    \small
    \caption{Training hyperparameters used for GCN on EvTouch-Objects.\label{tabel:appxtabel:gcn}}
    \label{tab:argparse_config}
    \begin{tabularx}{\linewidth}{Xc}
    \toprule
    \textbf{Hyperparameter} & \textbf{Value} \\
    \midrule
    Batch Size & 1 \\
    Total Train Epoch & 200 \\
    Number of Layers & 3 \\
    Learning Rate & 1e-3 \\
    Dropout Rate & 0.7 \\
    Maximum hop distance & 3 \\
    $L$ & 8 \\
    ANN Threshold $\vartheta$ & 8 \\
    $\alpha^*$ & -0.25 \\
    $\beta^*$ & 1 \\
    \bottomrule
\end{tabularx}
\end{table}
\begin{table}[htb]
    \centering
    \small
    \caption{Training hyperparameters used for ViT-S experiments on CIFAR-10, CIFAR-100, ImageNet, CIFAR10-DVS , and N-Caltech101. \label{tabel:appxtabel:vit-s}}
    \label{tab:training_config}
    \begin{tabularx}{\linewidth}{Xc}
    \toprule
    \textbf{Hyperparameter} & \textbf{Value} \\
    \midrule
    Train Epoch except ImageNet & 300 \\
    ImageNet Train Epoch & 100 \\
    Warmup Epoch & 5 \\
    CIFAR-10/100, ImageNet Initial Learning Rate & 1.5e-4 \\
    CIFAR10-DVS Initial Learning Rate & 8.09e-5 \\
    N-Caltech101 Initial Learning Rate & 2.25e-4 \\
    Weight Decay & 0.05 \\
    Batch Size & 64 \\
    Patch Size & 16 \\
    MLP ratio & 4 \\
    Rate of stochastic depth & 0.1 \\
    Label Smoothing Factor & 0.1 \\
    Cutmix & 1.0 \\
    Mixup & 1 \\
    Mixup probability & 0.5 \\
    Random Erasing (RE) probability & 0.25 \\
    RE Max erasing area & 0.4 \\
    RE Aspect of erasing area & 0.3 \\
    repeated augmentation(RA) & 3 \\
    Data AutoAugment & True \\
    LR Scheduler & Cosine LR \\
    augmentation policies for training & False \\
    $L$ & 16 \\
    $\alpha^*$ & $-1/2$ \\
    $\beta^*$ & $7/16$ \\
    Locality Self-Attention & False \\
    Shifted Patch Tokenization & False \\

    \bottomrule
    \end{tabularx}
\end{table}

\subsection{Experimental Setup}


\textbf{Datasets.} We conduct experiments on three widely used static benchmarks: CIFAR-10, CIFAR-100~\cite{krizhevsky2009learning}, and ImageNet~\cite{5206848}. CIFAR-10 comprises 60,000 color images across 10 classes, with 6,000 images per class. The dataset is divided into 50,000 training images and 10,000 test images. CIFAR-100 has the same number of images but spans 100 classes, with 600 images per class. CIFAR-100 follows the same split as CIFAR-10, with 50,000 training images and 10,000 test images. ImageNet contains over 1.28 million images for training and 50,000 images in the validation set.

We also used three dynamic neuromorphic datasets: CIFAR10-DVS~\cite{Li2017CIFAR10DVSAE}, N-Caltech101~\cite{li_andreeto_ranzato_perona_2022} and EvTouch-Objects\cite{gu2020tactilesgnet}. CIFAR10-DVS converts the static CIFAR-10 dataset into event streams. It was recorded using a DVS128 sensor capturing the LCD monitor displaying moving images from the original dataset. The dataset maintains the same 10 classes as CIFAR-10. N-Caltech101 serves as the neuromorphic counterpart to the Caltech101 dataset. It was generated by mounting an ATIS sensor on a pan-tilt unit to perform saccadic movements while viewing static images. EvTouch-Objects is an event-based tactile object recognition dataset collected using an event-driven tactile sensor during physical interactions with 36 objects.


\begin{table*}[t]
    \centering
    \caption{Performance on CIFAR-10 and CIFAR-100 with VGG-16 and ResNet-20 architectures. We compare our method with existing state-of-the-art approaches. The method marked with * is obtained via direct training.}
    \label{tab:v16r20-performance}
    \begin{tabular}{lccccc|ccc|ccc|ccc}
        \hline
        \multirow{3}{*}{\textbf{Method}} & \multirow{3}{*}{\textbf{Polarity}}& \multirow{3}{*}{\textbf{Multi-Spike}}& \multicolumn{6}{c|}{\textbf{CIFAR-10}} & \multicolumn{6}{c}{\textbf{CIFAR-100}} \\
        \cline{4-15}
        & & &\multicolumn{3}{c|}{\textbf{VGG-16}} & \multicolumn{3}{c|}{\textbf{ResNet-20}} & \multicolumn{3}{c|}{\textbf{VGG-16}} & \multicolumn{3}{c}{\textbf{ResNet-20}} \\
        \cline{4-15}
        & & &\textbf{ANN} & \textbf{T} & \textbf{Acc} & \textbf{ANN}  & \textbf{T} & \textbf{Acc} & \textbf{ANN} & \textbf{T} & \textbf{Acc} & \textbf{ANN} & \textbf{T} & \textbf{Acc} \\
        \hline

        RMP-L*~\cite{Guo_Liu_Chen_Zhang_Peng_Zhang_Huang_Ma_2023}& × & ×& - & 4 & 93.33 & - & 4 & 91.89 & - & 4 & 72.55 & - & 4 & 66.65 \\
        TransferL*~\cite{10843312}  & ×   & ×& - & 8 & 91.55 & - & - & - & - & 8 & 64.79 & -     & - & - \\

        OPI~\cite{Bu_Ding_Yu_Huang_2022}     & ×   & \checkmark         & 94.57 & 8 & 90.96 & 92.74 & 8 & 66.24 & 76.31 & 8 & 60.49 & 70.43 & 8 & 23.09 \\
        Off-LTL~\cite{Yang_Wu_Zhang_Chua_Wang_Li_2022}& × & ×& 94.05 & 16 & 93.04 & 95.36 & 16 & 94.82 & 74.42 & 16 & 74.19 & 76.36 & 16 & 76.12 \\
        On-LTL~\cite{Yang_Wu_Zhang_Chua_Wang_Li_2022}& × & ×& 94.05 & 16 & 92.85 & 95.36 & 16 & 91.33 & 74.42 & 16 & 71.09 & 76.36 & 16 & 73.22 \\
        SRP~\cite{Hao_Bu_Ding_Huang_Yu_2023}   & ×    & \checkmark      & 95.52 & 8 & 95.52 & 91.77 & 8 & 91.37 & 76.28 & 8 & 76.25 & 69.94 & 8 & 62.94 \\
        
        QCFS~\cite{Bu_Fang_Ding_Dai_Yu_Huang_2022}  & ×  & \checkmark   & 95.52 & 1 & 88.41 & 91.77 & 1 & 62.43 & 76.28 & 8 & 73.96 & 69.94 & 8 & 55.37 \\
        SlipReLU~\cite{Jiang_Anumasa_Masi_Xiong_Gu_2023}& × & ×& 93.82 & 1 & 88.17 & 82.07 & 1 & 80.99 & 68.46& 1 & 64.21 & 50.79 & 1 & 48.12 \\
        COS~\cite{Hao_Ding_Bu_Huang_Yu_2023}   & ×     & \checkmark     & 95.51 & 1 & 94.90 & 91.77 & 1 & 89.88 & 76.28& 1 & 74.24 & 69.97 & 1 & 59.22 \\
        EMORE~\cite{huang2024converting}        & ×    & \checkmark     & 95.21 & 1 & 88.46 & 85.18 & 1 & 65.99 & 74.86& 1 & 62.27 & 62.34 & 1 & 21.78 \\
        ASG~\cite{10843312}          & ×   & ×    & 91.56 & 8 & 91.55 & - & - & - & 64.81& 8 & 64.79 & - & - & - \\
        DNI-SNM~\cite{Wang_Liu_Zhang_Luo_Qu_2024} & \checkmark & × & 95.67 & 8 & 95.14 & - & - & - & 77.29 & 8 & 76.35 & - & - & - \\

        \multirow{1}{*}{\textbf{Ours}}& \checkmark & \checkmark& \multirow{1}{*}{\textbf{95.67}} & \textbf{1} & \textbf{95.67} & \multirow{1}{*}{\textbf{93.78}} & \textbf{1} & \textbf{93.78} & \multirow{1}{*}{\textbf{76.71}} & \textbf{1} & \textbf{76.71} & \multirow{1}{*}{\textbf{69.91}} & \textbf{1} & \textbf{69.91} \\
        \hline
    \end{tabular}
\end{table*}



\textbf{Network Architectures.} We consider four representative architectures: VGG-16~\cite{simonyan2014very}, ResNet-20~\cite{he2016deep}, Graph Convolution Network(GCN)~\cite{gu2020tactilesgnet} and ViT-S~\cite{dosovitskiy2020image}. For ViT-S, the attention modules involve softmax operations that are not directly compatible with spike-based computation. We therefore follow the SNN-friendly attention conversion scheme proposed in~\cite{You_Xu_Nie_Deng_Guo_Wang_He_2024}.

\textbf{Training Settings.} We adopt a training strategy in which theoretical hyperparameter initialization precedes network optimization. Specifically, the quantization configuration is determined prior to training. For fair comparison with existing ANN-to-SNN conversion methods, we follow the setting in~\cite{Bu_Fang_Ding_Dai_Yu_Huang_2022} and fix the number of quantization levels as $L=8$. Based on the theoretical analysis in Section~\ref{section:3:1}, the quantization threshold is initialized as $\vartheta_0=8$. A grid search is then performed over the $\alpha$-$\beta$ space to identify a feasible parameter configuration satisfying the entropy-preservation criterion. This process is conducted offline as a one-time analysis. During network training, $\vartheta$ is jointly optimized with network weights through backpropagation, while $L$ remains fixed.

Once the quantization configuration is determined, different training strategies are adopted for different architectures. Due to the simple structure of convolutional networks, VGG-16, ResNet-20 and GCN can be trained directly using the proposed PQA. In contrast, transformer architectures such as ViT-S exhibit more complex computational structures, involving attention mechanisms and global feature interactions. Therefore, ViT-S is first pretrained with standard ReLU activations and then fine-tuned with PQA using the optimized $(\alpha^*, \beta^*, L, \vartheta_0)$.

The detailed training hyperparameters used for VGG-16 and ResNet-20 are listed in Table~\ref{tabel:appxtabel:vgg_and_resnet}. The corresponding hyperparameter settings for the GCN model are provided separately in Table~\ref{tabel:appxtabel:gcn}. For the ViT-S, which does not include batch normalization (BN) layers, we follow the training settings of SpikeZIP-TF\cite{You_Xu_Nie_Deng_Guo_Wang_He_2024}. These settings are consistently applied across both the large-scale static and dynamic image datasets, with detailed hyperparameters summarized in Table~\ref{tabel:appxtabel:vit-s}.

\subsection{Performance Evaluation}\label{experiment:acc_compare}
\textit{1) PMSM Achieves State-of-the-Art Accuracy on CNN Architectures at the First Timestep:} We first evaluate the proposed conversion framework on VGG-16 and ResNet-20 architectures, with results summarized in Table~\ref{tab:v16r20-performance}. Our method consistently outperforms prior state-of-the-art conversion approaches across all settings. In particular, under the single-timestep setting ($T=1$), PMSM achieves clear improvements over the recent COS framework\cite{Hao_Ding_Bu_Huang_Yu_2023} on both CIFAR-10 and CIFAR-100 datasets. Specifically, the proposed method improves the Top-1 accuracy by 0.77\% and 3.90\% on VGG-16 and ResNet-20 for CIFAR-10, respectively. More notably, on the more challenging CIFAR-100 dataset, the accuracy gains further increase to 2.47\% and 10.69\%. Furthermore, on CIFAR-100 with VGG-16, PMSM achieves 76.71\% Top-1 accuracy at the first timestep, surpassing SRP~\cite{Hao_Bu_Ding_Huang_Yu_2023}, which requires 8 timesteps to attain 76.25\% accuracy. For a more comprehensive comparison, we further include state-of-the-art direct training methods (marked with *) as baselines. The proposed method consistently achieves higher accuracy under equal or substantially lower latency. For instance, on CIFAR-100, our method reaches 76.71\% at $T=1$, compared to 72.55\% achieved by RMP-L~\cite{Guo_Liu_Chen_Zhang_Peng_Zhang_Huang_Ma_2023} at $T=4$. These results demonstrate the effectiveness of the proposed PQA-based conversion framework in maintaining high representation fidelity under ultra-low-latency inference.
\begin{table}[!ht]
    \centering
    \caption{
    Comparison of ANN-to-SNN conversion, direct training, and our proposed PMSM methods using the ViT-S architecture. Methods marked with * are direct training methods.}
    \label{tab:vit-s-performance}
    \begin{tabularx}{\linewidth}{Xccccc}
        \hline
        \multirow{2}{*}{\textbf{Method}}   &    \multirow{2}{*}{\textbf{Polarity}}  &    \multirow{2}{*}{\textbf{Multi-Spike}}       &           \multicolumn{3}{c}{\textbf{CIFAR-10}}  \\ \cline{4-6}
        &  &  & \textbf{ANN}                   & \textbf{T}  & \textbf{Acc} \\
        \hline

        SpikFormer*~\cite{zhou2023spikformer}        & ×     & ×          & -                              & 4           & 95.51          \\
        Sdformer*~\cite{yao2023spike}          & ×        & ×                                 & -                              & 4           & 95.60          \\

        SpikedAttention~\cite{Hwang_Lee_Park_Lee_Kung_2024}    & ×     & ×                                     & 97.50                           & 24          & 97.30          \\
        MST*~\cite{Wang_Fang_Cao_Zhang_Wang_Xu_2023}   & ×            & ×                                        & 98.14                           & 256         & 97.27          \\
        SpikeZIP-TF~\cite{You_Xu_Nie_Deng_Guo_Wang_He_2024}  & \checkmark     & ×                                          & 99.20                           & 16          & 97.70          \\
        \multirow{1}{*}{\textbf{Ours}}&     \checkmark     & \checkmark         & \multirow{1}{*}{\textbf{98.85}} & \textbf{1} & \textbf{98.62} \\
        
        \hline
        \multirow{2}{*}{\textbf{Method}}   &    \multirow{2}{*}{\textbf{Polarity}}  &    \multirow{2}{*}{\textbf{Multi-Spike}}       &           \multicolumn{3}{c}{\textbf{CIFAR-100}}  \\ \cline{4-6}
        &  &  & \textbf{ANN}                   & \textbf{T}  & \textbf{Acc} \\
        \hline

        SpikFormer*~\cite{zhou2023spikformer}       & ×   & ×           & -                              & 4           & 78.21          \\
        Sdformer*~\cite{yao2023spike}                      & ×      & ×                     & -                              & 4           & 78.40          \\

        SpikedAttention~\cite{Hwang_Lee_Park_Lee_Kung_2024}                & ×       & ×             & 87.70                           & 24          & 86.30          \\
        MST*~\cite{Wang_Fang_Cao_Zhang_Wang_Xu_2023}                             & ×   & ×                        & 88.72                           & 256         & 86.91          \\
        SpikeZIP-TF~\cite{You_Xu_Nie_Deng_Guo_Wang_He_2024}                    & \checkmark     & ×                     & 91.90                           & 16          & 87.30          \\
        \multirow{1}{*}{\textbf{Ours}} &            \checkmark     & \checkmark           & \multirow{1}{*}{\textbf{90.56}} & \textbf{1}  & \textbf{89.56} \\

        \hline
         \multirow{2}{*}{\textbf{Method}}   &    \multirow{2}{*}{\textbf{Polarity}}  &    \multirow{2}{*}{\textbf{Multi-Spike}}       &           \multicolumn{3}{c}{\textbf{ImageNet}}  \\ \cline{4-6}
        &  &  & \textbf{ANN}                   & \textbf{T}  & \textbf{Acc} \\
        \hline
        SpikFormer*~\cite{zhou2023spikformer}                    & × & ×  & -                              & 4           & 74.81         \\
        Sdformer*~\cite{yao2023spike}                      &      ×   & ×                    & -                              & 4           & 77.07         \\

        SpikedAttention~\cite{Hwang_Lee_Park_Lee_Kung_2024}                &       ×      & ×                & 79.30                           & 48          & 77.20          \\
        MST*~\cite{Wang_Fang_Cao_Zhang_Wang_Xu_2023}                            &   ×    & ×                      & 80.51                           & 512         & 78.51         \\
        SpikeZIP-TF~\cite{You_Xu_Nie_Deng_Guo_Wang_He_2024}                    &    \checkmark   & ×                      & 82.34                           & 64          & 81.45          \\
        \multirow{1}{*}{\textbf{Ours}} &            \checkmark    & \checkmark            & \multirow{1}{*}{\textbf{82.34}} & \textbf{1}  & \textbf{81.61} \\
 
        \hline
    \end{tabularx}
\end{table}

\textit{2) PMSM Extends Effectively to Transformer Architectures Under Single-Timestep Constraints:} We then extend the evaluation to the transformer-based ViT-S architecture, with results reported in Table~\ref{tab:vit-s-performance}. The proposed method again achieves state-of-the-art performance, particularly at the first timestep. On ImageNet, it reaches 81.61\% Top-1 accuracy at $T=1$, a performance level not previously achieved at such latency. A slight accuracy gap is observed between the converted SNNs and its ANNs counterpart, which is mainly attributed to the use of a baseline SNN-friendly attention conversion scheme~\cite{You_Xu_Nie_Deng_Guo_Wang_He_2024} that typically requires multiple timesteps to approximate ANN outputs. Despite this limitation, the proposed method still achieves the best accuracy with the lowest latency, demonstrating its effectiveness even without dedicated optimization for attention modules.

\begin{table}[!ht]
    \centering
    \caption{Accuracy comparison of the ViT-S model against other methods on the CIFAR10-DVS and N-Caltech101 dataset. Methods marked with * are direct training methods.}
    \begin{tabularx}{\linewidth}{Xccccc}
    \hline
          \multirow{2}{*}{\textbf{Method}}   &    \multirow{2}{*}{\textbf{Polarity}}  &    \multirow{2}{*}{\textbf{Multi-Spike}}       &           \multicolumn{3}{c}{\textbf{CIFAR-10-DVS}}  \\ \cline{4-6}
        &  &  & \textbf{ANN}                   & \textbf{T}  & \textbf{Acc} \\
        \hline
        tdBN*~\cite{zheng2021going}&×&×& - & 10 & 67.80 \\ 
        Sdformer*~\cite{yao2023spike}&× &×& - & 16 & 80.00 \\ 
        Sglformer*~\cite{zhou2023enhancing}& ×&×& - & 16 & 82.60 \\ 
        MST*~\cite{Wang_Fang_Cao_Zhang_Wang_Xu_2023}&× &×& - & 512 & 88.10 \\ 
        OST*~\cite{Song_Song_Xiao_Sun_2024} &×&×& - & 1 & 81.20 \\ 
        AT-VGG9*~\cite{10947245}&× &×& - & 20 & 87.62 \\ 
        \multirow{1}{*}{SpikeZIP-TF~\cite{You_Xu_Nie_Deng_Guo_Wang_He_2024}}&\checkmark &×& \multirow{1}{*}{90.40} & 32 & 87.60 \\ 
        \multirow{1}{*}{\textbf{Ours}}&\checkmark&\checkmark & \multirow{1}{*}{\textbf{90.20}} & \textbf{1} & \textbf{89.43} \\ 
        \hline
        \multirow{2}{*}{\textbf{Method}}   &    \multirow{2}{*}{\textbf{Polarity}}  &    \multirow{2}{*}{\textbf{Multi-Spike}}       &           \multicolumn{3}{c}{\textbf{N-Caltech101}}  \\ \cline{4-6}
        &  &  & \textbf{ANN}                   & \textbf{T}  & \textbf{Acc} \\
        \hline
        Sdformer*~\cite{yao2023spike}&×&× & - & 16 & 81.80 \\ 
        SpikingReformer*~\cite{shi2024spikingresformer}&×&× & - & 16 & 81.29 \\ 
        QKFormer*~\cite{zhou2024qkformer}& ×&×& - & 16 & 83.58 \\ 
        STAtten*~\cite{Lee_2025_CVPR}&× &×& - & 16 & 84.25 \\ 
        SpikFormer*~\cite{zhou2023spikformer}&×&× & - & 5 & 72.83 \\ 
        STCA-SNN*~\cite{10.3389/fnins.2023.1261543}&× &×& - & 14 & 80.88 \\ 
        \multirow{1}{*}{\textbf{Ours}}&\checkmark&\checkmark & \multirow{1}{*}{\textbf{93.44}} & \textbf{1} & \textbf{92.58} \\
        \hline
    \end{tabularx}
    \label{tab:vits:cifar10dvs}
\end{table}  
\begin{table}[!ht]
    \centering
    \caption{Accuracy comparison of graph convolutional neural networks against other methods on the Ev-Objects tactile event dataset. Methods marked with * are direct training methods.}
    \begin{tabularx}{\linewidth}{Xccccc}
    \hline
          \multirow{2}{*}{\textbf{Method}}   &    \multirow{2}{*}{\textbf{Polarity}}  &    \multirow{2}{*}{\textbf{Multi-Spike}}       &           \multicolumn{3}{c}{\textbf{EvTouch-Objects}}  \\ \cline{4-6}
        &  &  & \textbf{ANN}                   & \textbf{T}  & \textbf{Acc} \\
        \hline
        TactileGCN*~\cite{garcia2019tactilegcn}&×&×& - & 250 & 66.70 \\ 
        SLAYER*~\cite{shrestha2018slayer}&× &×& - & 250 & 81.40 \\ 
        ResNet-18*~\cite{guo2025event}&× &×& - & 250 & 81.10 \\ 
        Tactile-SNN*~\cite{xu2022recurrent}&× &×& - & 250 & 75.40 \\ 
        GCN*~\cite{hamilton2017inductive}&× &×& - & 250 & 73.70 \\ 
        Grid CNN*~\cite{gu2020tactilesgnet}& ×&×& - & 250 & 88.40 \\ 
        MLP*~\cite{gu2020tactilesgnet}&× &×& - & 250 & 85.97 \\ 
        TactileSGNet*~\cite{gu2020tactilesgnet} &×&×& - & 250 & 89.44 \\ 
        
        \multirow{1}{*}{\textbf{Ours}}&\checkmark&\checkmark & \multirow{1}{*}{\textbf{90.97}} & \textbf{1} & \textbf{90.97} \\ 
        \hline
    \end{tabularx}
    \label{tab:gcn:EvTouch-Objects}
\end{table}
\textit{3) PMSM Demonstrates Strong Generalization Across Event-Based Transformer and GCN Architectures:} To further assess the generalization capability of PMSM across event-based Transformer and GCN architectures, we conduct experiments on three event-driven benchmarks, including CIFAR10-DVS, N-Caltech101, and EvTouch-Objects. As shown in Table~\ref{tab:vits:cifar10dvs} and Table~\ref{tab:gcn:EvTouch-Objects}, the proposed method achieves 89.43\% accuracy on CIFAR10-DVS at $T=1$, outperforming prior methods by at least 1.3\% while reducing latency by an order of magnitude ($10\times$). On the N-Caltech101 dataset, the proposed method achieves 92.58\% accuracy at $T=1$. To the best of our knowledge, this is the first report of high-fidelity ANN-to-SNN conversion at single-timestep latency on this dataset. Furthermore, it surpasses state-of-the-art direct training approaches by more than 8\%, demonstrating its ability to effectively transfer high-precision ANN representations to SNNs without requiring extended temporal inference. On the EvTouch-Objects tactile event dataset, the proposed method achieves 90.97\% accuracy at a single timestep, surpassing the previous best-performing model, TactileSGNet\cite{gu2020tactilesgnet}, by 1.53\% while reducing the inference latency from 250 timesteps to only one timestep. Notably, unlike prior methods that rely on direct SNN training, our approach attains state-of-the-art performance through ANN-to-SNN conversion under ultra-low-latency constraints. These results further demonstrate that the proposed framework generalizes effectively beyond conventional CNN and Transformer architectures to graph-based tactile perception models.


\begin{figure}[!ht]
    \centering
    \includegraphics[width=\linewidth]{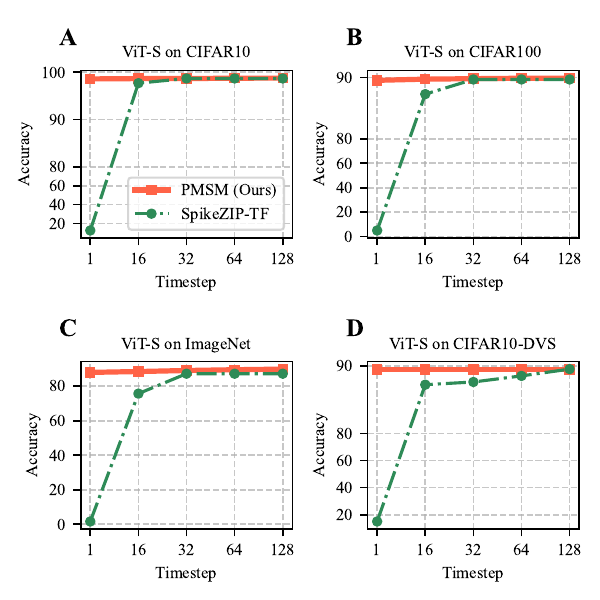}
    \caption{Accuracy comparison across different inference timesteps ($T=1$ to $128$) between the proposed method and SpikeZIP-TF\cite{You_Xu_Nie_Deng_Guo_Wang_He_2024} on the ViT-S architecture across multiple datasets. \textbf{(A)} CIFAR-10 dataset. \textbf{(B)} CIFAR-100 dataset. \textbf{(C)} ImageNet dataset. \textbf{(D)} CIFAR10-DVS event-based dataset.
\label{fig:fig-time-and-acc}
    }
\end{figure}
\textit{4) PMSM Achieves Stable Performance Across Multiple Timesteps:} Furthermore, we analyze the impact of inference timesteps on model performance by evaluating ViT-S across a wide range of timesteps on CIFAR-10, CIFAR-100, ImageNet, and CIFAR10-DVS, as illustrated in Fig.~\ref{fig:fig-time-and-acc}. The results consistently show that, beyond achieving ultra-low-latency inference, our framework can effectively leverages the temporal dynamics to deliver stable performance, indicating that the proposed PMSM does not sacrifice long-term temporal expressiveness for low-latency inference. In comparison, the recently proposed SpikeZIP-TF\cite{You_Xu_Nie_Deng_Guo_Wang_He_2024} exhibits significant performance degradation under the single-timestep setting. Specifically, on static image datasets, SpikeZIP-TF typically requires $T=32$ to achieve nearly lossless conversion performance, while substantially larger timesteps are further required on dynamic event-based datasets to maintain competitive accuracy. In contrast, the proposed PMSM achieves high-fidelity conversion already at $T=1$ while still maintaining stable performance as the timestep increases.

\subsection{Layer-wise Entropy Ratio Across Different Architectures}
\begin{figure}[htb]
    \centering
    \includegraphics[width=\linewidth]{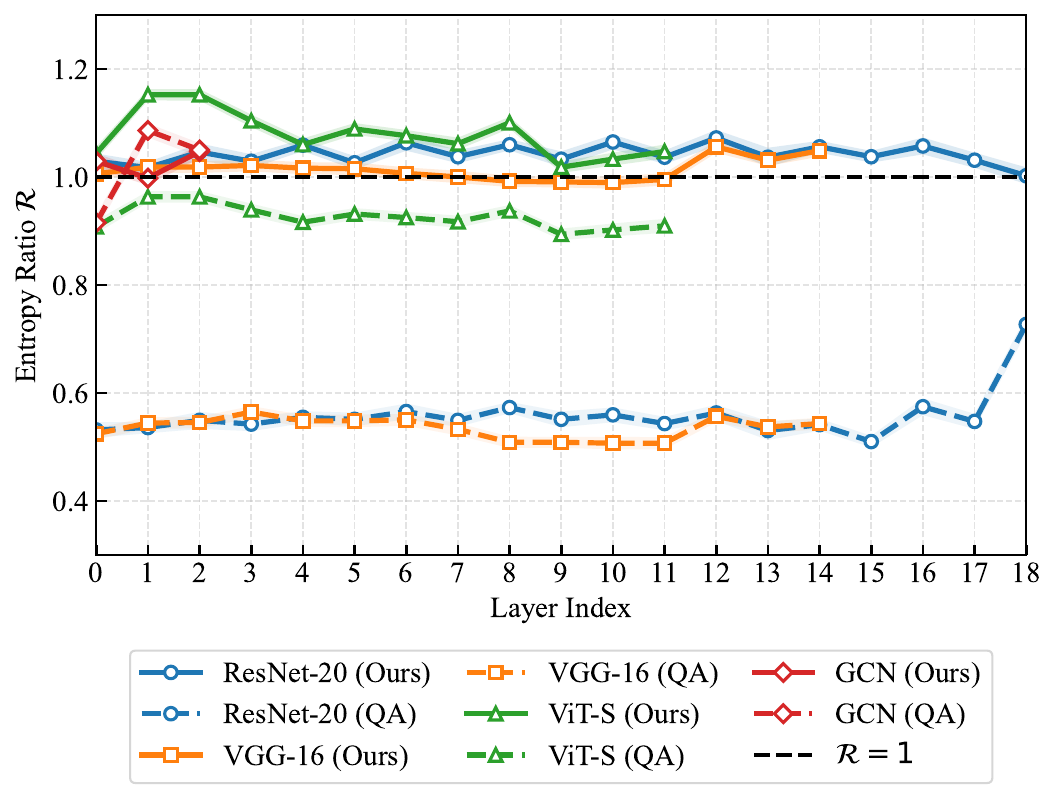}
    \caption{Layer-wise entropy ratio $\mathcal{R}$ computed using the learned quantization thresholds after training for ResNet-20, VGG-16, ViT-S, and GCN. The dashed line denotes the lossless-information-preservation boundary ($\mathcal{R}=1$). Compared with QA\cite{Bu_Fang_Ding_Dai_Yu_Huang_2022}, the proposed PMSM consistently maintains entropy ratios close to the theoretical lossless region across different architectures.
    \label{fig:entropy_ratio}}
\end{figure}
To further investigate the information-preservation characteristics of the learned quantization parameters, we compute the entropy ratio $\mathcal{R}$ for each layer after training across different architectures, including ResNet-20, VGG-16, ViT-S, and GCN. The results are presented in Fig.~\ref{fig:entropy_ratio}.

Several observations can be drawn. First, PMSM consistently maintains entropy ratios close to or above the lossless-information-preservation boundary ($\mathcal{R}=1$) across all architectures, whereas the conventional QA-based quantization scheme generally exhibits lower entropy ratios. This result indicates that the proposed entropy-guided initialization strategy provides a favorable starting point for optimization and enables the learned quantization parameters to remain within an information-preserving regime even after training. Among the CNN-based architectures, the difference between PMSM and QA is particularly pronounced. For both ResNet-20 and VGG-16, the entropy ratios of QA remain around 0.5--0.6 throughout most layers, indicating substantial information loss during quantization. In contrast, PMSM maintains entropy ratios close to unity across the entire network depth. In particular, VGG-16 exhibits entropy ratios almost perfectly aligned with the theoretical lossless-information-preservation condition, while ResNet-20 consistently operates within the near-lossless regime despite the increased representational complexity introduced by residual connections. These observations suggest that PMSM effectively preserves activation information that would otherwise be discarded by conventional quantization.

Among the architectures equipped with PMSM, ViT-S exhibits the highest entropy ratios, with most layers maintaining values between approximately 1.03 and 1.15. According to the entropy analysis, entropy ratios slightly larger than one indicate the presence of moderate representational redundancy beyond the original activation information. However, Vision Transformers employ self-attention mechanisms that dynamically reweight token interactions and selectively focus on informative features. Consequently, the additional representational redundancy introduced by quantization does not necessarily degrade performance and may instead provide richer feature representations for subsequent attention computation. This observation is consistent with the superior conversion performance achieved by PMSM on both static and event-based Transformer benchmarks. Furthermore, although GCN differs fundamentally from CNNs and Transformers by operating on graph-structured tactile events, both PMSM and QA maintain entropy ratios close to unity across all layers. Despite the relatively small entropy gap between the two methods, PMSM still achieves superior ANN-to-SNN conversion performance on the EvTouch-Objects dataset. This result suggests that the benefits of PMSM are not limited to image-based architectures and can generalize effectively to graph-based tactile perception models.

Overall, despite the substantial differences in architecture and optimization dynamics among ResNet-20, VGG-16, ViT-S, and GCN, the corresponding PMSM-based models consistently converge to entropy-preserving operating regimes after training. These findings provide empirical evidence that the proposed hyperparameter initialization strategy reliably guides optimization toward information-preserving quantization configurations and generalizes effectively across CNN, Transformer, and graph-based architectures.

\begin{figure*}[!ht]
    \centering
    \includegraphics[width=\linewidth]{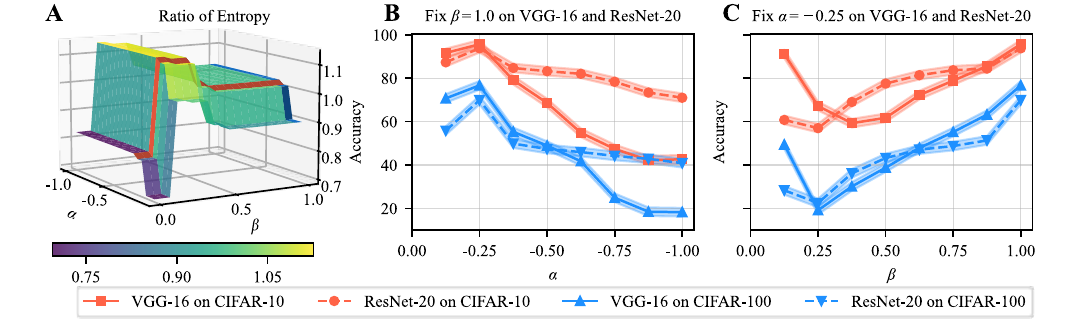}
    \caption{Entropy-guided analysis of quantization boundaries and corresponding accuracy variation under different $\alpha$ and $\beta$ configurations on VGG-16 and ResNet-20 architectures for CIFAR-10 and CIFAR-100 datasets. \textbf{(A)} Entropy ratio $\mathcal{R}$ under varying $\alpha$ and $\beta$. \textbf{(B)} Accuracy when $\beta$ is fixed at $1.0$ and $\alpha$ varies. \textbf{(C)} Accuracy when $\alpha$ is fixed at $-0.25$ and $\beta$ varies.
    \label{fig:fig4}}
\end{figure*}
\subsection{Validation of Hyperparameter Initialization Strategy}  \label{experiment:ablation}

To systematically evaluate the effectiveness of the proposed hyperparameter initialization strategy, we conduct a dedicated analysis on the impact of the quantization boundaries $\alpha$ and $\beta$ across different architectures and datasets. Specifically, experiments are performed on VGG-16~\cite{simonyan2014very} and ResNet-20~\cite{he2016deep} using CIFAR-10 and CIFAR-100. With the quantization configuration initialized as
$L$ = 8 and $\vartheta$ = 8, we perform a grid-based evaluation over the $\alpha$-$\beta$ parameter space to examine how different boundary settings affect the entropy ratio $\mathcal{R} = \frac{\mathcal{H}_{\text{PQA}}}{\mathcal{H}_{\text{BN}}}$. 

As shown in Fig.~\ref{fig:fig4}(A), when fixing $\beta = 1.0$ and gradually decreasing $\alpha$ from $-0.125$ to $-1$, the entropy ratio $\mathcal{R}$ exhibits three distinct regimes. Specifically, when $\mathcal{R}<1$, the quantization range is insufficient to capture the underlying activation distribution, resulting in entropy loss and forming the \emph{information-loss region}. As $\alpha$ decreases further, $\mathcal{R}$ approaches unity ($\mathcal{R}\approx1$), indicating that the entropy of the quantized activations closely matches that of the original activations. This defines the \emph{near-lossless region}, where information preservation is maximized. However, when the quantization range continues to expand, the entropy ratio exceeds one ($\mathcal{R}>1$), suggesting that additional quantization states introduce redundant uncertainty beyond the original activation distribution. This leads to a \emph{noise-dominated region}, where excessive range expansion increases quantization noise without contributing useful information. This behavior is consistent with the corresponding accuracy trends in Fig.~\ref{fig:fig4}(B), where performance first improves as information preservation increases, and then degrades due to noise accumulation. A similar pattern is observed when fixing $\alpha$ and increasing $\beta$ from $0.125$ to $1$, as illustrated in Fig.~\ref{fig:fig4}(C), further confirming the effect of quantization boundaries on information preservation.

Overall, these results validate that the entropy ratio $\mathcal{R}$ provides an effective criterion for guiding the selection of quantization boundaries. Moreover, the consistent trends observed across different architectures and datasets indicate that the proposed entropy-guided strategy is  generalizable.
\subsection{Energy Consumption Analysis}  \label{experiment:power_consume}


To quantitatively evaluate the energy efficiency of our method, we estimate the computational cost of SNNs relative to their ANNs counterparts by calculating the number of operations, following the widely adopted methodology in prior work~\cite{zhou2023spikformer,yao2023spike, 10032591}. The computational cost of ANNs is measured by floating-point operations (FLOPs), whereas the cost of SNNs is dominated by synaptic operations (SOPs). The energy consumption of ANNs for processing a single image can be formulated as:
\begin{equation}
    E_{\mathrm{ANN}} = E_{\mathrm{MAC}} \times \mathrm{FLOPs},
\end{equation}
where $E_{\mathrm{ANN}}$ denotes the energy consumption of the ANNs model. We assume that FLOPs are implemented on 45nm hardware\cite{horowitz20141}, where $E_{\mathrm{MAC}} = 4.6\,\mathrm{pJ}$. For SNNs, the computational cost is determined by both FLOPs in the first layer and the number of synaptic operations (SOPs) in subsequent layers:
\begin{equation}
\begin{aligned} 
    E_{\mathrm{SNN}} 
    &= E_{\mathrm{MAC}} \times \mathrm{FLOPs}(l=1)  \\
    &+ E_{\mathrm{AC}} \times \left(\sum_{a=2}^{A} SOPs_{\mathrm{Conv}}^{a} + \sum_{b=1}^{B} SOPs_{\mathrm{FC}}^{b}\right),  
\end{aligned}
\end{equation}
where $E_{\mathrm{SNN}}$ denotes the energy consumption of the SNNs model. $\mathrm{FLOPs}(l=1)$ denotes the first layer that encodes static RGB images into spike sequences. $SOPs_{\mathrm{Conv}}^{a}$ and $SOPs_{\mathrm{FC}}^{b}$ denote the number of SOPs in the $a$-th convolutional layer and the $b$-th fully connected layer, respectively. We assume that SOPs are implemented on the same 45nm hardware\cite{horowitz20141}, where $E_{\mathrm{AC}} = 0.9\,\mathrm{pJ}$. The number of SOPs at layer $l$ is computed as:
\begin{equation}
    SOPs(l) = fr \times T \times \mathrm{FLOPs}(l),
\end{equation}
where $l$ denotes the $l$-th layer in SNNs, $fr$ is the firing rate, and $T$ is the number of timesteps.


\begin{table}[!ht]
    \centering

    \caption{Power consumption comparison between PMSM and conventional QA-based ANN-to-SNN conversion methods under different inference timesteps. The results include VGG-16 on CIFAR-10 and CIFAR-100, and GCN on the EvTouch-Objects tactile event dataset. FLOPs and SOPs denote floating-point operations and spiking operations, respectively.}
    
    \begin{tabular}{lccccr}
    \hline
        \textbf{Method}  & \textbf{Acc} & \textbf{$T$} & \textbf{FLOPs(M)} & \textbf{SOPs(M)} & \textbf{Energy($\mu$J)} \\ \hline
        \multicolumn{6}{c}{CIFAR-10} \\ \hline
        ANN~\cite{10361844} & 95.73  & - & 332.07 & 0 & 1527.20 (Base)  \\ 
        QA~\cite{Bu_Fang_Ding_Dai_Yu_Huang_2022}  & 88.41 & 1 & 1.77 & 27.69 &  $33.06\,(\downarrow \!46\times)$  \\ 
        \textbf{Ours}  & 95.67 & 1 & 1.77  & 6.78 & $\mathbf{14.24\,(\downarrow \!107\times)}$  \\ \hdashline
        QA~\cite{Bu_Fang_Ding_Dai_Yu_Huang_2022} & 91.18 & 2 & 3.54  & 61.52 & $71.65\,(\downarrow \!21\times)$  \\ 
        \textbf{Ours}  & 95.67 & 2 & 3.54  & 13.62 & $\mathbf{28.54\,(\downarrow \!54\times)}$  \\ \hline
        \multicolumn{6}{c}{CIFAR-100} \\ \hline
        ANN~\cite{10361844} &  77.22  & - & 332.07 & 0 & 1527.20 (Base)  \\ 
        QA~\cite{Bu_Fang_Ding_Dai_Yu_Huang_2022} & 47.05 & 1 & 1.77  & 27.93 & $33.28\,(\downarrow \!46\times)$  \\ 
        \textbf{Ours}& 76.71 & 1 & 1.77  & 6.50 & $\mathbf{13.99\,(\downarrow \!109\times)}$   \\ \hdashline
        QA~\cite{Bu_Fang_Ding_Dai_Yu_Huang_2022}  & 57.56 & 2 & 3.54  & 62.97 & $72.96\,(\downarrow \!21\times)$  \\ 
        \textbf{Ours}  & 76.71 & 2 & 3.54  & 13.04 & $\mathbf{28.02\,(\downarrow \!55\times)}$  \\ \hline
        \multicolumn{6}{c}{EvTouch-Objects} \\ \hline
        QA~\cite{Bu_Fang_Ding_Dai_Yu_Huang_2022} & 2.78 & 1 & -  & - & -  \\ 
        QA~\cite{Bu_Fang_Ding_Dai_Yu_Huang_2022} & 86.11 & 64 & -  & 0.89 & 0.80 (Base)   \\ 
        \textbf{Ours}& 90.97 & 1 & - & 0.41 & $\mathbf{0.37\,(\downarrow \!2\times)}$   \\  \hline
    \end{tabular}
    \label{tabel:power_compare}
\end{table}


The detailed FLOPs, SOPs, and corresponding energy consumption under different inference timesteps are reported in Table~\ref{tabel:power_compare} for both the VGG-16 architecture on CIFAR-10/CIFAR-100 and the GCN architecture on the EvTouch-Objects dataset. Compared with the corresponding ANNs\cite{10361844}, our method achieves substantial energy savings. At $T=1$, the energy consumption is reduced by 107$\times$ on CIFAR-10 and 109$\times$ on CIFAR-100. At $T=2$, the reductions remain significant at 54$\times$ and 55$\times$, respectively. Furthermore, compared with SNNs obtained using conventional quantized activation (QA) functions~\cite{Bu_Fang_Ding_Dai_Yu_Huang_2022}, our method consistently exhibits significantly lower energy consumption. Specifically, on CIFAR-10, the proposed method reduces energy consumption by 56.93\% and 60.17\% at $T=1$ and $T=2$, respectively, while further improving the classification accuracy by 7.26\% and 4.49\%. Similarly, on CIFAR-100, our method achieves energy reductions of 57.96\% at $T=1$ and 61.60\% at $T=2$, accompanied by additional accuracy gains of 29.66\% and 19.15\%, respectively. 
Consistent trends are also observed on the EvTouch-Objects tactile event dataset using a GCN architecture. Notably, the conventional QA-based conversion method fails to achieve a valid ANN-to-SNN conversion at $T=1$. This limitation is particularly pronounced for event-driven tactile data, where the input events are inherently sparse and a single timestep cannot accumulate sufficient spike information to faithfully represent the quantized activations. As a result, the converted SNN collapses at ultra-low latency. In contrast, PMSM enhances the information capacity of each timestep by allowing neurons to emit multiple polarity spikes within a single inference step, thereby avoiding the need for temporal accumulation. As a result, PMSM achieves 90.97\% accuracy with only a single inference timestep. Even when compared with the QA-based counterpart operating at $T=64$, PMSM improves the recognition accuracy by 4.86\% while reducing the energy consumption by 53.75\%. This result indicates that the proposed PMSM not only benefits CNN-based image classification models, but also effectively transfers to graph-based tactile perception networks, achieving superior accuracy-energy trade-offs under ultra-low-latency inference.
\begin{figure}[htb]
    \centering
    \includegraphics[width=\linewidth]{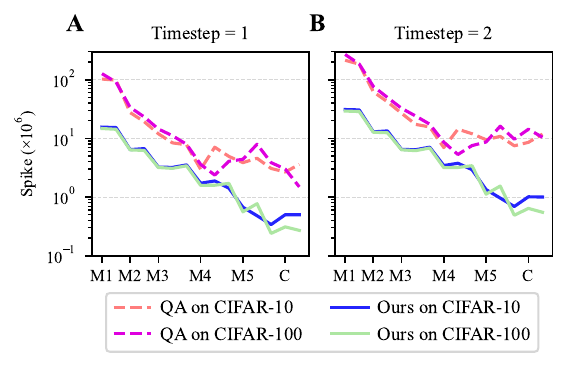} 
    \caption{Comparison of layer-wise spike counts between the proposed method and QA~\cite{Bu_Fang_Ding_Dai_Yu_Huang_2022} on the VGG-16 architecture. \textbf{(A)} Results at $T=1$. \textbf{(B)} Results at $T=2$. The x-axis denotes neuron layer labels, where M$x$-N$y$ and C-N$y$ represent the $y$-th layer in Module-$x$ and the classifier module of VGG-16, respectively.}
    \label{fig:fig5}
\end{figure}

We further analyze the layer-wise spike activity across the network to better understand the efficiency of the proposed method. As shown in Fig.~\ref{fig:fig5}, the proposed method consistently produces fewer spikes per layer compared to baseline approaches~\cite{Bu_Fang_Ding_Dai_Yu_Huang_2022}. Notably, this reduction in spike activity does not degrade model performance, but instead indicates a more efficient representation enabled by the proposed quantization and neuron design. As a result, the overall energy consumption is significantly reduced, demonstrating the advantage of our method in achieving both high accuracy and energy efficiency.




\section{Discussion}


While our proposed conversion framework achieves accurate and low-latency ANN-to-SNN mapping through the use of both multiple spikes and polarity, it is also designed with practical deployment in mind. Specifically, our method is hardware-friendly and can be adapted to both existing and emerging neuromorphic platforms with minimal modification.

IBM TrueNorth~\cite{doi:10.1126science.1254642} is a digital neuromorphic chip, where each spike is transmitted as a 1-bit event. In such systems, representing negative spikes would require separate pathways for positive and negative signals. In contrast, SpiNNaker~\cite{6750072} supports 32-bit event packets, enabling each spike to carry additional information such as polarity and count. This makes it possible to realize our multi-spike and polarity-based mapping through software-defined encoding. Similarly, Intel Loihi 2~\cite{9605018} supports custom neuron models, signed synaptic weights, and multi-bit spike signaling, making it well-suited for the direct and efficient implementation of our method.

Importantly, the theoretical framework developed in this work reveals properties that are not only compatible with current hardware but also valuable for guiding the design of future neuromorphic chips. The ability to achieve nearly lossless ANN-to-SNN conversion within a single timestep, while improving performance across timesteps, presents a compelling target for hardware–software co-design.

\section{Conclusion}  \label{section:Conclusion}

This paper presents an ANN-to-SNN conversion framework targeting high accuracy and low latency, incorporating a Polarity Quantized Activation (PQA) function and Augmented Integrate-and-Fire (AIF) neurons to faithfully approximate the activation values of ANNs at the first timestep.
The PQA function adopts a tailored hyperparameter initialization strategy to preserve the activation distribution under quantized representation, thereby enabling lossless information transfer in the quantization domain and serving as a solid foundation for accurate and efficient conversion.
AIF neurons emit multiple spikes with polarity by leveraging both positive and negative thresholds, allowing SNNs to approximate quantized activations at the first timestep, thus reducing inference latency while maintaining accuracy. We further provide a theoretical analysis of the activation shift, revealing a membrane potential modulation effect induced by the temporal dynamics of SNNs. Specifically, we prove that with extremely few timesteps ($T = 1$), the conversion error is approximately zero. Experimental results on both image classification and event-based benchmarks, including CIFAR-10, CIFAR-100, ImageNet, CIFAR10-DVS, N-Caltech101, and EvTouch-Objects, demonstrate that PMSM surpasses existing ANN-to-SNN conversion methods in accuracy without requiring fine-tuning or additional training. Moreover, PMSM consistently matches or exceeds state-of-the-art direct training methods while maintaining ultra-low-latency inference and strong generalization across diverse architectures, including CNNs, Vision Transformers, and graph convolutional networks. Although PMSM shows competitive performance on both static and event-driven recognition tasks, its applicability to more complex problems such as object detection and semantic segmentation remains to be explored. Overall, this work represents a promising step toward accurate and efficient ANN-to-SNN conversion with ultra-low latency.




\bibliographystyle{IEEEtran}

\bibliography{03_references_tpami_cleaned}

\end{document}